\documentclass[preprint,12pt,authoryear]{elsarticle}

\usepackage{amssymb}
\usepackage{amsthm}

\usepackage{enumitem}
\usepackage{amsmath}
\usepackage{makecell}
\usepackage{multirow}
\usepackage{adjustbox}
\usepackage{booktabs}
\usepackage{graphicx}
\usepackage{float}
\usepackage{xcolor}
\usepackage{mathtools}
\usepackage{etoolbox}
\usepackage{setspace}
\usepackage{units}
\usepackage{subcaption}
\usepackage{comment}
\usepackage[hidelinks]{hyperref}

\newtheorem*{assumption*}{\assumptionnumber}
\providecommand{\assumptionnumber}{}
\makeatletter
\newenvironment{assumption}[2]
 {%
  \renewcommand{\assumptionnumber}{Assumption #1: #2}%
  \begin{assumption*}%
  \protected@edef\@currentlabel{#1}%
 }
 {%
  \end{assumption*}
 }

\DeclareMathOperator*{\argmax}{arg\,max}

\DeclarePairedDelimiter\floor{\lfloor}{\rfloor}

\definecolor{red(ncs)}{rgb}{0.0, 0.0, 0.0}

\newcommand{\rd}[1]{\textcolor{red(ncs)}{#1}}

\journal{}

\begin{document}

\begin{frontmatter}



\title{Spatial-temporal recurrent reinforcement learning for autonomous ships}


\author[label1]{Martin Waltz\corref{CorrespondingAuthor}}
\ead{martin.waltz@tu-dresden.de}

\author[label1,label2]{Ostap Okhrin}

\address[label1]{Technische Universität Dresden, Chair of Econometrics and Statistics, esp. in the Transport Sector, Dresden, 01062, Germany}
\address[label2]{Center for Scalable Data Analytics and Artificial Intelligence (ScaDS.AI), Dresden/Leipzig, Germany}

\cortext[CorrespondingAuthor]{Corresponding author}
\begin{abstract}
\rd{This} paper proposes a spatial-temporal recurrent neural network architecture for \rd{deep $Q$-networks that can be used to} steer an autonomous ship. The network design \rd{makes it possible to handle} an arbitrary number of surrounding target ships while offering robustness to partial observability. \rd{Furthermore}, a state-of-the-art collision risk metric is proposed to enable an easier assessment of different situations by the agent. The COLREG rules of maritime traffic are explicitly considered in the design of the reward function. The final policy is validated on a custom set of newly created single-ship encounters called `Around the Clock' problems and the commonly \rd{used} Imazu (1987) problems, which include 18 multi-ship scenarios. \rd{Performance comparisons with artificial potential field and velocity obstacle methods demonstrate the potential of the proposed approach for maritime path planning. Furthermore, the new architecture exhibits robustness when it is deployed in multi-agent scenarios and it is} compatible with other deep reinforcement learning algorithms, including actor-critic frameworks. 
\end{abstract}




\end{frontmatter}



\section{Introduction}
\rd{The safety-critical traffic domain can greatly benefit from the use of reliable, autonomously controlled ships. Despite efforts to improve safety measures, human error continues to be the primary cause of maritime accidents. The Annual Overview of Marine Casualties and Incidents 2021 by the \cite{EMSA21} revealed that over 53\% of maritime accidents between 2014 and 2020 were caused by human actions. This highlights the potential for autonomous surface vehicles (ASVs) to significantly reduce accident rates in maritime operations while also improving the energy and time efficiency, extending operational reliability and precision, and increasing flexibility for dangerous missions \citep{liu2016unmanned}. As a result, there has been a surge of interest, both academic and industrial, in designing ASVs, with projects such as \cite{Rolls-Royce}, \cite{Autoship}, and \cite{YaraBirkeland} being carried out alongside studies like those of  \cite{johansen2016ship},  \cite{lyu2018fast}, \cite{cheng2018concise}, \cite{zhao2019colregs}, and \cite{hart2022vessel}.}

\rd{A crucial characteristic of an ASV is the ability to reliably plan and follow a path toward a specified goal position. The own ship (OS), which is the controlled ASV, should thereby perform collision avoidance (COLAV) with surrounding vessels, called target ships (TSs). Path planning is a well-established field that originated from robotics (see \citealp[Chapter 7]{siciliano2008springer}) and there exists a plethora of algorithms developed for this purpose. The most prominent approaches include the artificial potential field (APF) method \citep{APF_original}, velocity obstacles (VO) \citep{fiorini1998motion}, genetic algorithms \citep{holland1992adaptation}, and sampling-based algorithms \citep{kuffner2000rrt}. We review these approaches in more detail in Section \ref{sec:literature_review}. It is important to note that in addition to the inherent limitations of existing algorithms, the \emph{Convention on the International Regulations for Preventing Collisions at Sea} (COLREG) is often overlooked. This convention is a set of rules established by the \cite{COLREGs1972} to enhance maritime traffic safety by defining the proper behaviour for seafarers in certain encounter situations. All vessels in high seas or connected navigable waters must comply with the COLREG rules, making it imperative that any practical path-planning algorithm takes them into account. However, current industry-standard autopilots cannot handle the complex requirements of high-level path-planning and COLREG-compliant COLAV. In fact, according to \cite{heiberg2022risk}, developing a dynamics model and control law that can simultaneously perform path following and COLREG-compliant COLAV is an infeasible task for traditional control methods.}

\rd{In order to overcome these limitations, there have been recent proposals to apply the latest advancements in reinforcement learning (RL; \citealp{sutton2018reinforcement}) to the domain of maritime operations. RL is a subfield of artificial intelligence in which an agent, such as a controlled vessel in this case, learns to maximise a reward signal through trial-and-error interaction with its environment. \cite{silver2021reward} even hypothesised that any form of intelligence and its associated capabilities can be thought of  as maximising a reward signal. Crucially, RL relies on approximations to solve multistage decision problems that could be solved with dynamic programming but have a computationally intractable solution. Hence, some authors refer to RL as \emph{approximate dynamic programming}, and the field has a strong relationship with control theory \citep{bertsekas2019reinforcement}. Recent contributions leveraging the relationship between RL and control theory include those of \cite{vrabie2009adaptive}, who proposed an adaptive optimal control method for continuous-time linear systems based on policy iteration, and \cite{xin2022online}, who outlined an online RL algorithm for solving multiplayer non-zero sum games.}

\rd{In recent years, RL has been combined with deep learning (DL; \citealp{lecun2015deep}), which uses neural networks to approximate arbitrary functions with high accuracy \citep{matsuo2022deep}.  This intersection of RL and DL is known as deep reinforcement learning (DRL), which has achieved remarkable performance in various applications, including complex strategy games \citep{vinyals2019grandmaster}, molecule optimisation \citep{zhou2019optimization}, advanced racing simulations \citep{wurman2022outracing}, and even the autonomous navigation of stratospheric balloons over the Pacific Ocean \citep{bellemare2020autonomous}. The methodological basis of these works is often the deep $Q$-network (DQN) of \cite{mnih2015human}, which pairs the off-policy $Q$-learning algorithm of \cite{watkins1992q} with deep neural networks and paved the way for the recent success of DRL. Numerous modifications of the DQN have since been proposed by researchers such as \cite{van2016deep}, \cite{hessel2018rainbow}, \cite{d2021gaussian}, and \cite{waltz2022two}.}

\rd{There have been several proposals for DRL in the field of ASVs. For instance, \cite{cheng2018concise} presented a concise DRL algorithm for obstacle avoidance based on a DQN, although their study only considered static obstacles. \cite{xu2022path} modified the deep deterministic policy gradient (DDPG) algorithm of \cite{lillicrap2015continuous} to construct an autonomous COLAV algorithm that considered COLREGs. Nonetheless, their study only tested relatively simple scenarios. The DDPG algorithm, which is similar to the DQN but was designed for continuous action spaces, was also used by \cite{zhou2022obstacle} to develop an ASV obstacle avoidance method, but traffic rules were not explicitly considered. \cite{sawada2021automatic} implemented an automatic COLAV system based on the on-policy proximal policy optimisation algorithm \citep{schulman2017proximal}. The authors explicitly provided predictions of future vessel collisions via the obstacle zone target. \cite{li2021path} combined the DQN with the conventional APF method to design a COLAV algorithm. However, their neural network directly provides a heading change for the OS; therefore, the low-level control routine is not considered part of the DRL task. \cite{fan2022novel} constructed a DRL-based maritime COLAV algorithm based on the dueling DQN \citep{wang2016dueling}, although the validation again considered only simple scenarios with at most two target ships. Other notable contributions include the works of \cite{shen2019automatic}, \cite{guo2020autonomous}, \cite{xu2020intelligent}, \cite{meyer2020taming}, \cite{chun2021deep}, and \cite{xu2022colregs} on intelligent ASV control and COLAV using DRL.}

All the studies above have in common that the OS must aggregate the information of surrounding target ships to assess the collision risk of a situation and ultimately select an action. The TS information \rd{is delivered by the automatic identification system} (AIS), which is mandatory equipment for vessels of specific sizes \citep{lin2006comparison}. \rd{It} can be considered a feature vector for each TS that contains information such as \rd{the} position, course, and speed. On this basis, two critical challenges arise.

\rd{The first challenge is that the number of target ships can vary, making it difficult to process the input vectors with a fully connected neural network with a fixed input size. To address this challenge, various simplifying assumptions and practices have been used in the literature. For instance, \cite{chun2021deep} and \cite{xu2022path} considered only the target ship with the highest collision risk while ignoring all other information, which is not practical since multi-ship encounters require the consideration of multiple vessels. Other researchers such as \cite{xu2020intelligent} attempted to fix a certain number of target ships and train an extra agent for each configuration, which is also not practical since a different control policy is required for each possible number of ships. \cite{zhao2019colregs} clustered the input based on possible COLREG encounter situations (see subsection \ref{subsec:COLREGs}), but this approach has limitations when multiple target ships are in the same situation. Alternatively, some studies use different observation formats such as LiDAR beams \citep{li2021path, zhou2022obstacle, meyer2020taming} or visual grid input \citep{woo2020collision, sawada2021automatic}, without utilising the available AIS data.}

\rd{The second challenge in processing AIS data for COLAV is the creation of a robust framework under partial observability, considering that the received data may include delays, noise, and weather-based disturbances \citep{almalioglu2022deep}. AIS data may not have a high enough frequency to generate sufficient situational awareness in crowded areas \citep{heiberg2022risk}, which suggests the need for the algorithm to process information from several time steps instead of solely relying on the current observation. This approach can improve the ability of the algorithm to handle complex and dynamic situations, reducing the impact of noise and disturbances.}

In this paper, we leverage the potential of DRL by designing a simulation-based ASV agent that tackles both challenges. Our contributions to the literature are as follows:
\begin{itemize}
    \item A \emph{spatial-temporal recurrent} neural network architecture for the DQN is proposed. The approach extends \rd{the} prior work of \cite{everett2018motion, everett2021collision} from the robotic domain and \rd{makes it possible} to process information in multi-ship encounter situations with a variable number of target ships. \rd{Due} to the integration of the temporal recurrent component, we achieve robustness against partial observability. \rd{Furthermore, our approach offers a fast computational time and makes real-life implementation feasible.}
    \item A \emph{novel collision risk metric} based on the concept of the closest point of approach (CPA; \citealp{lenart1983collision}) and the ship domain \citep{goodwin1975statistical} is designed. Through the neural architecture and the collision risk metric, we avoid the \rd{common} hierarchical decomposition of the agent into different sub-controllers\rd{, as, for example, in \cite{johansen2016ship} and \cite{zhai2022intelligent}. Thus, we provide a robust end-to-end solution from AIS data to rudder angle control in maritime operations.}
    \item A suitable training environment with a COLREG-dependent spawning routine of target ships is designed. Additionally, we create \emph{Around the Clock} problems as a comprehensive test-bed \rd{for} single-ship encounters. \rd{Furthermore,} our agent is thoroughly validated in multi-encounter scenarios using the \emph{Imazu problems} \citep{sawada2021automatic} and two multi-agent situations called the \emph{Star problems}, \rd{which are} inspired by \cite{zhao2019colregs}.
    \item \rd{We compare our approach to two \emph{competitive benchmarks} in the form of the APF method of \cite{huang2019generalized} and the VO algorithm of \cite{kuwata2013safe}.} In all cases, \rd{our} agent successfully reaches a specified destination, safely avoids collisions with other vessels, obeys maritime traffic rules, and performs realistic steering actions. \rd{These findings highlight the potential of DRL as a promising alternative to traditional algorithmic approaches in the domain of ASV control.}
\end{itemize}

This paper is structured as follows: \rd{Section \ref{sec:literature_review} provides maritime background information and further references.} Section \ref{sec:model_maritime_traffic} introduces the modelling of maritime traffic, including the environmental dynamics model, details \rd{concerning} the COLREGs, and the new collision risk metric. Section \ref{sec:RL_methods} presents the RL methodology, outlines the proposed neural network architecture, and describes the state, action, and reward configuration. Section \ref{sec:training} discusses \rd{the} simulation environment used for training, while the validation scenarios are shown in Section \ref{sec:validation}. Section \ref{sec:conclusion} concludes \rd{this paper}. We \rd{have made} the source code \rd{for} this paper publicly available at \url{https://github.com/MarWaltz/TUD_RL} to enable full reproducibility.

\def\thesection{\color{red(ncs)}\arabic{section}}

\section{\rd{Maritime background}}\label{sec:literature_review}
\subsection{\rd{Guidance, navigation, and control}}
\rd{\cite{vagale2021path} defined ASVs as vessels that have the ability to operate independently without human guidance, navigation, and control (GNC). However, developing such a system is a complex task that requires the consideration of numerous aspects. At the core of an ASV is the navigation module, which estimates the vessel's current state based on various sensors such as inertial measurement units, global positioning systems, cameras, LiDAR sensors, or RADAR sensors \citep{liu2016unmanned}. Moreover, AIS messages from surrounding target ships can be processed. Using the navigation module's output, the guidance module of the GNC framework is responsible for generating a reliable geometric path in the form of a set of waypoints that the vessel needs to follow. The ASV system should take into account restrictions on the waterway, assess the collision risk, and perform COLAV with other vessels during this phase. Finally, the control module determines the necessary control forces required to follow the path set by the guidance system. Path following is traditionally achieved using methods such as line-of-sight (LOS) \citep{fossen2021handbook} or vector field guidance \citep{nelson2006vector} to generate a course command, which is then translated into a low-level control command using traditional approaches such as a proportional-integral-derivative controller. Recently, more advanced learning-based or control-theoretic approaches have also been considered for the latter step \citep{liu2018ship, woo2019deep, sandeepkumar2022unified, paulig2023robust}.}
\sloppy
\subsection{\rd{Practical challenges}}
\rd{Fully autonomous ships present a range of practical challenges that need to be addressed. First, environmental disturbances such as winds, waves, or currents can significantly impact the manoeuvrability of the controlled vessel \citep{fossen2021handbook,almalioglu2022deep} and may affect how vessels behave in encountered situations \citep{zhou2019review}. Additionally, controlling real ASV systems requires uncertainties resulting from unmodelled dynamics, underactuation, and faults in sensors, actuators, or communication links to be addressed \citep{liu2016unmanned}. Fault diagnosis is a crucial research area that is essential for ensuring the reliable operation of real-life systems \citep{gao2015survey,tao2023unsupervised}. Furthermore, the reliable transfer of learning algorithms from simulations to real-world scenarios is still an open research area, with current approaches being heavily dependent on the quality of the simulator \citep{ju2022transferring}. While approaches such as domain randomisation or domain adaptation may help bridge the sim-to-real gap \citep{zhao2020sim}, the maritime domain, with its various sources of disturbances, represents a challenging area for such methods. Finally, unexpected behaviour and the non-cooperation of other ships may arise in practical situations, and cybersecurity threats must also be taken into account \citep{akdaug2022collaborative}. For a more in-depth discussion of these practical issues, we refer readers to \cite{liu2016unmanned} and \cite{akdaug2022collaborative}.}

\subsection{\rd{Path planning and collision avoidance}}
\rd{While path following is mostly based on the minimisation of a cross-track or course error, the path-planning task involves strategic decision-making and includes COLAV with target ships. While some papers use a more detailed classification \citep{breivik2017mpc, eriksen2020hybrid}, the maritime literature generally distinguishes between two types of path planning: \emph{global} and \emph{local}. Although some algorithms can perform overlapping tasks \citep{vagale2021path}, global path planning focuses on finding a path from an initial state to a goal state while accounting for known obstacles. On the other hand, local path planning is reactive and aims to produce a safe and COLREG-compliant path based on online information, thereby enabling COLAV. The focus of this paper is on local path planning. In addition to the DRL-based algorithms outlined in the introduction, there are various conventional approaches for maritime path planning and COLAV, which we review in the following.}

\rd{A popular planning algorithm is the \emph{artificial potential field} (APF) method, which was first introduced by \cite{APF_original} and has since been adapted for the maritime domain \citep{lyu2018fast, lyu2019colregs, liu2023colregs}. The APF method defines an attractive field for the goal and repulsive fields for obstacles. By superimposing the resulting forces, the vessel is pushed towards the goal while avoiding collisions with other ships. The APF method requires a low computational effort and allows an efficient implementation in practice, but it has the known disadvantages of potentially trapping the vessel in local optima and making goals unreachable when obstacles are nearby \citep{ge2000new}.}

\rd{\emph{Velocity obstacles} (VO) represent another approach, which was first proposed by \cite{fiorini1998motion} and successfully adapted for the maritime domain by \cite{kuwata2013safe}. Instead of considering dynamically moving objects in a positional space, the VO method statically considers the space of velocities of nearby vessels. Based on this, the algorithm computes a velocity obstacle, which is a set of velocities that would lead to a collision in the future under the assumption of the linear movement of another vessel. Recently, non-linear and probabilistic trajectories have also been considered \citep{huang2018velocity, huang2019generalized}. After computing the VO for each vessel, the algorithm selects a velocity vector that does not lie in the union of the VO sets. The selection is performed via the optimisation of some objective, such as the deviation from a desired velocity. The disadvantage of VO approaches is that the solution space may be empty \citep{ribeiro2021velocity}, or the vessel may exhibit undesired oscillatory motions \citep{tang2023cooperative}.}

\rd{\emph{Genetic algorithms} (GAs) represent another widely used option for optimisation and particularly planning problems \citep{holland1992adaptation, ozturk2022review}. GAs use principles inspired by biological evolution, such as reproduction, mutation, and selection, to find the optimal solution to a problem. For instance, \cite{kim2017study} outlined an approach for autonomous maritime path planning based on a GA while specifying avoiding obstacles, reaching a target point, and minimising the travel time as the objective functions. Further recent marine GA contributions include those of \cite{ning2020colregs} and \cite{wang2021cooperative}. Although GAs have the potential to generate collision-free paths, they suffer from significant computation times, making them challenging to deploy in practice \citep{tam2010path}.}

\rd{Sampling-based algorithms like \emph{rapidly exploring random trees} (RRTs) represent more approaches from the robotics domain and are particularly useful for high-dimensional spaces and non-linear systems, where other methods can struggle \citep{kuffner2000rrt, lavalle2001randomized}. The basic idea behind RRTs is to build a tree structure that explores the search space in a randomised way, gradually expanding to cover more and more of the space. \cite{karaman2011sampling} provided a detailed analysis and proposed several competitive modifications of the original RRT algorithm. \cite{chiang2018colreg}, \cite{zaccone2019colreg}, and \cite{enevoldsen2021colregs} proposed different adaptions of RRT-based algorithms to the maritime domain that consider traffic rules. Sampling-based approaches have some primary drawbacks, including their potentially long computation times and their dependence on a forward simulator to predict the behaviour of target ships \citep{chiang2018colreg}.}

\rd{\emph{Model predictive control} (MPC) algorithms are a powerful class of algorithms that can be used for path planning. MPC algorithms can compute optimal trajectories using an environmental model that describes the surrounding environment in detail \citep{garcia1989model}. During optimisation, MPC can consider several factors that affect the trajectory, including nonlinear dynamics, constraints, and disturbances. \cite{johansen2016ship} proposed an MPC-based planner that controls the speed and course of a vessel while using LOS guidance and a proportional-integral controller to translate the course command into a rudder angle. Further maritime applications of MPC include those of \cite{abdelaal2018nonlinear}, \cite{hagen2018mpc}, and \cite{wang2020roboat}. The main limitations of MPC-based proposals when it comes to practical applications are possible convergence issues and the computational complexity \citep{hagen2018mpc}.}

\rd{This list of methods is far from being complete; in addition, there are several other algorithms for maritime path planning and COLAV, such as dynamic-window approaches \citep{serigstad2018hybrid}, ant colony optimisation \citep{lazarowska2015ship}, and fast marching methods \citep{liu2015path}. For further reading on this topic, we recommend the reviews provided by \cite{vagale2021}, \cite{zhai2022intelligent}, and \cite{ozturk2022review}. As mentioned above, those techniques are mainly used for path planning and rely on a separate module for path following that involves the low-level control of actuators. This contrasts with our DRL approach, which provides an end-to-end solution, tackling local path planning and following simultaneously.}

\section{Modelling of maritime traffic}\label{sec:model_maritime_traffic}
\subsection{Environmental dynamics}
We consider the \rd{Manoeuvring Modelling} Group (MMG) model of \cite{yasukawa2015introduction} to describe the dynamics of the full-scale KVLCC2 \citep{stern2011experience} tanker. Several related works \citep{cheng2018concise, heiberg2022risk, du2022colregs} instead rely on the miniature model ship CyberShip II\rd{,} with the hydrodynamic parameters identified by \cite{skjetne2004nonlinear}. However, we have chosen the KVLCC2 since it has a length between perpendiculars of $L_{pp} = \unit[320]{m}$, allowing for a more realistic setup on the ocean, and the corresponding dynamics model \rd{was} identified and extensively tested \rd{by} \cite{yasukawa2015introduction}. Our simulation relies on the following assumptions, which are frequently imposed when maritime traffic \rd{is modelled} \citep{meyer2020taming, heiberg2022risk}:

\begin{assumption}{1}{Calm sea}
There are no external disturbances in \rd{the} form of wind, waves, or currents.
\end{assumption}

\begin{assumption}{2}{Motion restriction}\label{assump:motion_restriction}
The vessel is located on a horizontal plane with no heave, pitch, and rolling motion.
\end{assumption}

\begin{figure}
    \centering
    \includegraphics[width=0.5\textwidth]{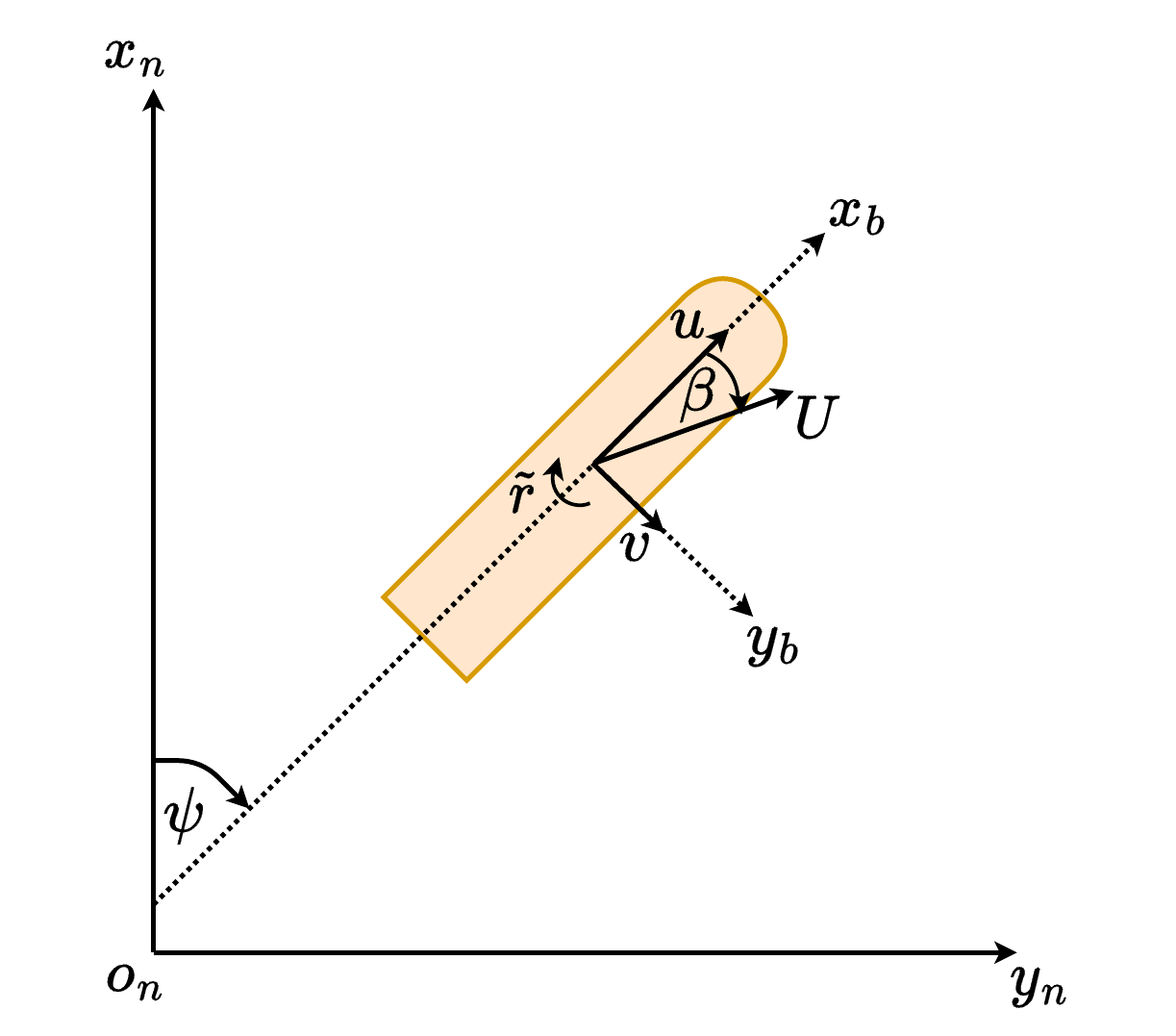}
    \caption{\rd{Visualisation} of the coordinate systems $\{n\}$ and $\{b\}$\rd{,} similar to \cite{yasukawa2015introduction}.}
    \label{fig:ship_coordinates}
\end{figure}

\noindent
Consequently, the model consists of 3 degrees of freedom\rd{, and} the navigational state of the vessel is described by \rd{$\eta = (x_n, y_n, \psi)^\top$}. The elements $x_n$ and $y_n$ are \rd{the} north and east coordinates relative to a coordinate origin $o_n$ in the \emph{North-East-Down} system $\{n\}$, where $\psi$ is the vessel's heading and \rd{is} defined as the angle between the $x_n$-axis and the $x_b$-axis of the \emph{body-fixed} reference frame $\{b\}$, which is centred at the midship position. In $\{b\}$, $x_b$ corresponds to the longitudinal axis, while $y_b$ is the transversal (starboard) axis. The velocity of the vessel is described by \rd{$\nu = (u, v, \Tilde{r})^\top$}, where $u$ and $v$ are the components in \rd{the} $x_b$ and $y_b$ \rd{directions}, respectively, and $\Tilde{r}$ is the yaw rate. Furthermore, we define the drift angle $\beta = \arctan(v/u)$, the total speed $U = \sqrt{u^2 + v^2}$, and the course angle $\chi = \psi + \beta$ \citep{fossen2021handbook}. Figure \ref{fig:ship_coordinates} illustrates the coordinate systems. The following set of equations connects $\{n\}$ and $\{b\}$ via rotation:
\begin{equation}\label{eq:MMG_model_1}
\begin{aligned}
    \dot{x}_n &= u \cos{\psi} - v \sin{\psi},\\
    \dot{y}_n &= u \sin{\psi} + v \cos{\psi},\\
    \dot{\psi} &= \Tilde{r},
\end{aligned}
\end{equation}
where $\dot{w}$ denotes the (later component-wise) first\rd{-}order derivative of $w$. The equations of motion governing the dynamics of the vessel are
\begin{equation}\label{eq:MMG_model_2}
\begin{aligned}
    (m + m_{x_b}) \dot{u} - (m + m_{y_b}) v \Tilde{r} - x_G m \Tilde{r}^2 &= X = X_H + X_R + X_P, \\
    (m + m_{y_b}) \dot{v} + (m + m_{x_b})u\Tilde{r} + x_G m \dot{\Tilde{r}} &= Y = Y_H + Y_R, \\
    (I_{zG} + x_{G}^{2} m + J_z) \dot{\Tilde{r}} + x_G m (\dot{v} + u\Tilde{r}) &= N_m = N_H + N_R,
\end{aligned}
\end{equation}
where $m$ is the ship's mass, $m_{x_b}$ \rd{and} $m_{y_b}$ are the added masses in \rd{the} $x_b$ and $y_b$ directions, respectively, $x_G$ is the longitudinal coordinate in $\{b\}$ of the \rd{centre} of gravity (COG) of \rd{the} ship, $I_{zG}$ is the moment of inertia of the ship around the COG, and $J_z$ is the added moment of inertia. \rd{Furthermore}, $X$ is the surge force, $Y$ is the lateral force, and $N_m$ is the yaw moment around midship. These forces consist of their respective hull (H), rudder (R), and propeller (P) components. \cite{yasukawa2015introduction} \rd{provided} detailed expressions for each force\rd{, along with} parameters for the full-scale KVLCC2 tanker, and we refer the reader interested in the fine details to this paper. Crucially, by adjusting the tanker's rudder angle $\delta$, we \rd{can} control the related rudder forces $X_R$, $Y_R$, and $N_R$ and \rd{realise} steering commands.

We use a subscript $t$ to refer to a particular quantity at time $t$, e.g. $\eta_t$ and $\nu_t$ \rd{refer to the} navigation and velocity vectors, respectively. We discretise the dynamics using a step size of \unit[3]{\rd{s}}\rd{, and} thus $\eta_{t+1}$ corresponds to \unit[3]{\rd{s}} in real time after $\eta_{t}$. To obtain $\eta_{t+1}$ and $\nu_{t+1}$, we solve the systems of equations (\ref{eq:MMG_model_1}) and (\ref{eq:MMG_model_2}) for the derivatives $\dot{\eta}_t$ and $\dot{\nu}_t$, and \rd{we} use the ballistic method of \cite{treiber2015comparing}. The latter consists of an Euler update for the speeds and a trapezoidal update for the positions. 

\subsection{International Regulations
for Preventing Collisions at Sea}\label{subsec:COLREGs}
\subsubsection{Overview}
Compliance with the COLREG rules of the \cite{COLREGs1972} is mandatory for vessels operating \rd{in} high seas. However, the 41 rules lack specific, measurable quantities \rd{that can be used to} determine whether a particular \rd{behaviour} was COLREG-compliant or not. Notably, in the 1970s, when the rules were published, there were no autonomous vessel systems, and the \rd{regulations were} tailored to human seafarers and decision-makers. As pointed out by \cite{heiberg2022risk}, modern \rd{optimisation} approaches do not necessarily produce COLREG-compliant actions. For example, \rd{COLAV} actions must be substantial and visible to make the agent's intentions transparent to other traffic participants. However, such \rd{behaviour mostly does not result in} fuel or time efficiency. The combination of autonomous and human-controlled vessels and potential changes to the regulatory rule set constitute an active area of research \citep{zhou2020study, miyoshi2022rules}.

\begin{figure}
    \centering \includegraphics[width=0.7\textwidth]{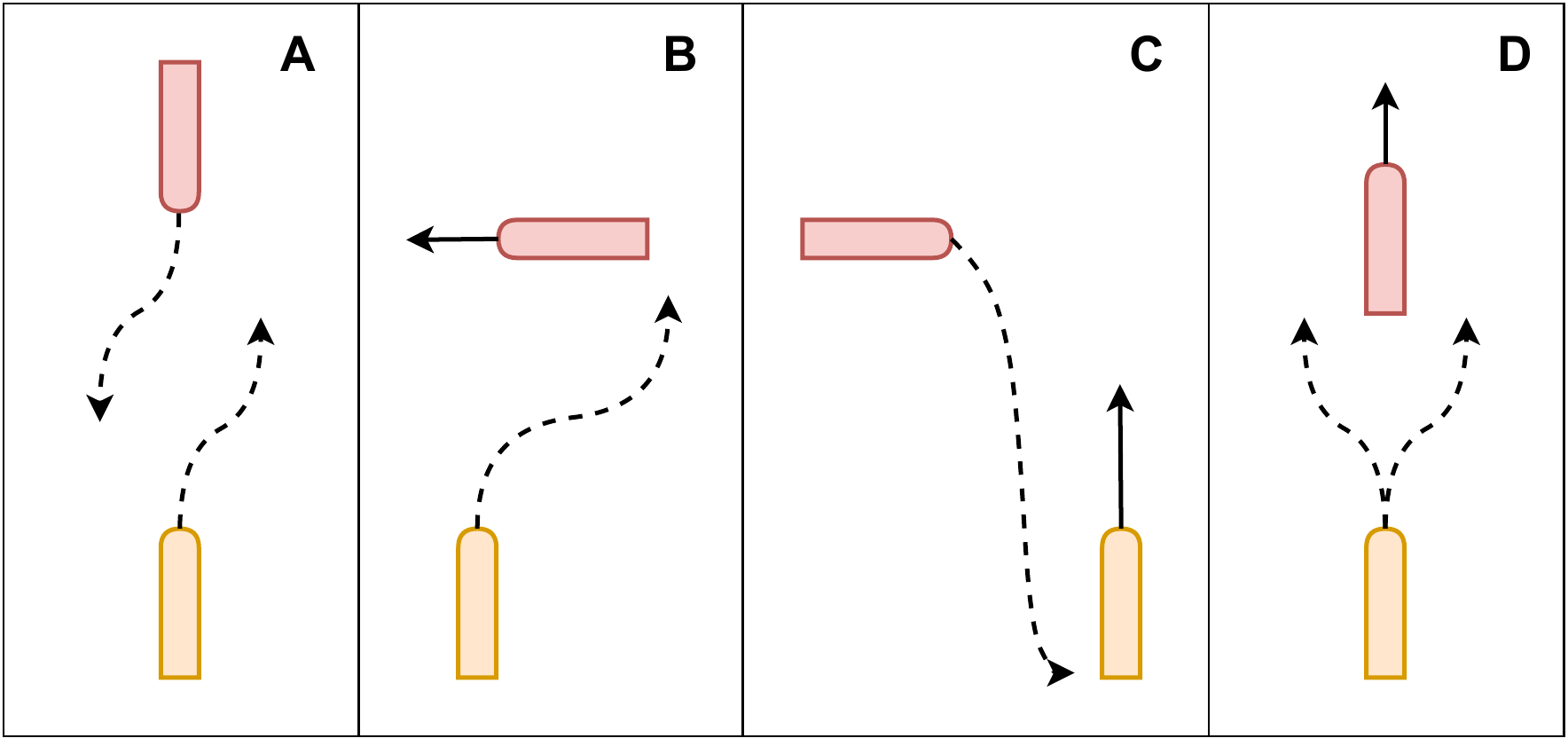}
    \caption{\rd{Visualisation} of the COLREG encounter \rd{situations} based on \cite{vagale2021path}. The OS is yellow and the TS is red. There are four scenarios: \rd{(A) head-on}, \rd{(B)} starboard crossing, \rd{(C)} port crossing, and \rd{(D)} overtaking. The give-way vessel is depicted with a dashed line, while the stand-on vessel \rd{is depicted with} a solid line.}
    \label{fig:COLREG_viz}
\end{figure}

For the reader's convenience, \ref{appendix:COLREGs} contains concrete passages of the rules that are especially relevant for maritime traffic \rd{modelling}. To briefly \rd{summarise} the most important COLREGs, each vessel should proceed at a safe speed to effectively avoid collisions, determine if the risk of collision exists, and take substantial \rd{COLAV} actions to pass other vessels at a safe distance. Moreover, the COLREGs categorise vessels into give-way and stand-on \rd{vessels}, depending on their role in one of four possible encounter situations \rd{(}see Figure \ref{fig:COLREG_viz}\rd{)}. Generally, give-way vessels are required to keep out of the other vessel's way, while stand-on vessels should keep their \rd{current} course and speed. For example, both vessels are classified as give-way in a \rd{head-on} scenario (Figure \ref{fig:COLREG_viz}, Panel \rd{A}) and should change their course to starboard. 

Similar to \cite{woo2020collision} and \cite{xu2022colregs}, we will assume that the target ships move \rd{linearly} and \rd{deterministically} in our simulation. Therefore, our work focuses on cases \rd{in which} the OS is the give-way ship and is responsible for actively avoiding collisions. \rd{Subsection} \ref{subsec:COLREG_scenarios} discusses the criteria for each encounter scenario.

\subsubsection{Scenario classification}\label{subsec:COLREG_scenarios}
\begin{figure}
    \centering
    \includegraphics[width=0.5\textwidth]{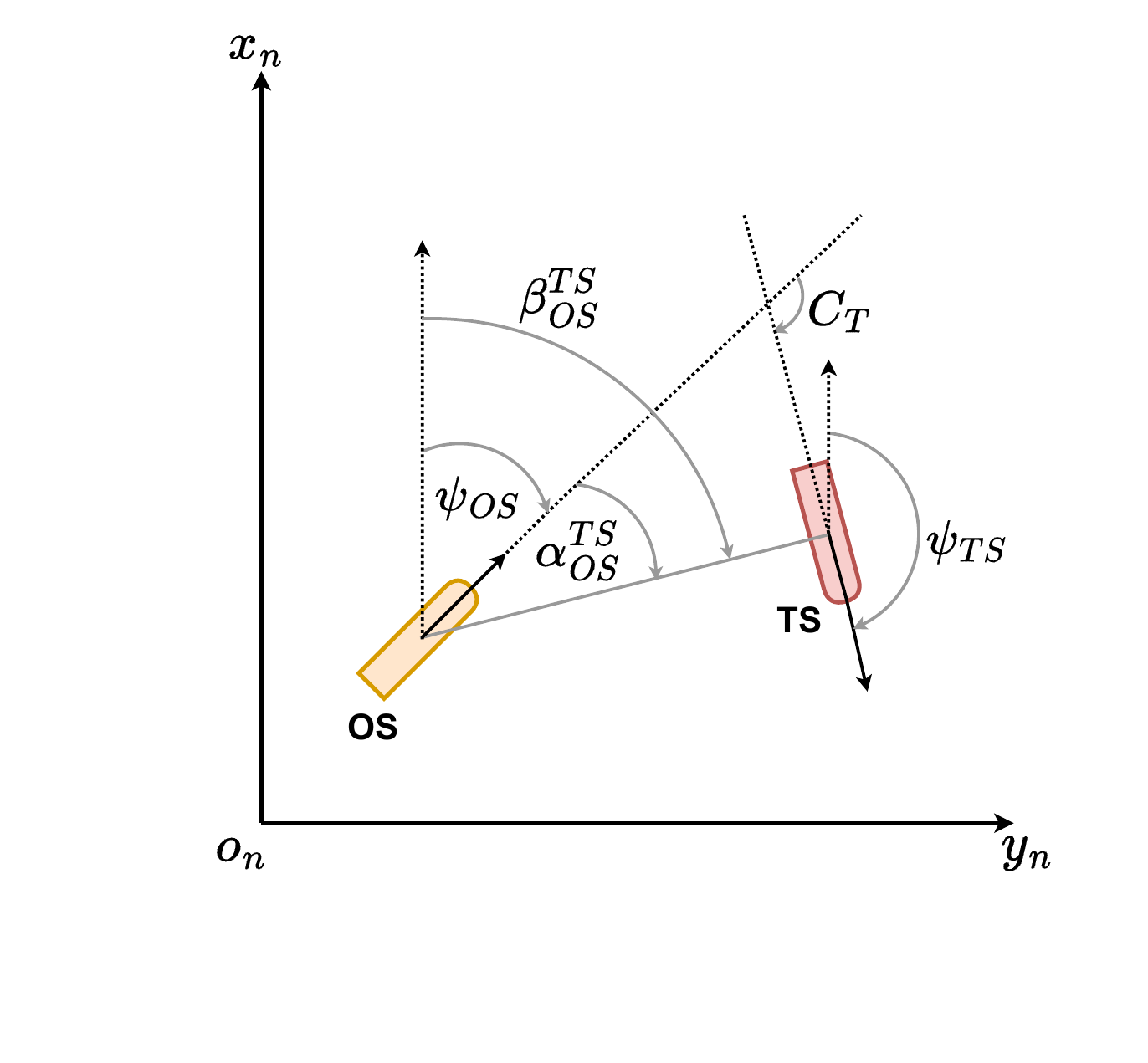}
    \caption{\rd{Visualisation} of the heading intersection angle $C_T$, the relative bearing $\alpha_{OS}^{TS}$, and the absolute bearing $\beta_{OS}^{TS}$.}
    \label{fig:OS_TS_intersection_bearing}
\end{figure}

Whether a scenario at sea is classified as one of the four encounter situations depends on the two vessels' angular and positional constellation, \rd{which is} measured \rd{using} two quantities \rd{(}see Figure \ref{fig:OS_TS_intersection_bearing}\rd{)}. \rd{The first is} the heading intersection angle $C_T = \left[\psi_{TS} - \psi_{OS}\right]^{2\pi}_0$, where $\psi_{TS}$ and $\psi_{OS}$ are the headings of the TS and the OS, respectively, with the clipping operation to $[a, a+ 2\pi)$ for $a \in \{-\pi, 0\}$ defined as $[\cdot]^{a+2\pi}_a$:

\[ \left[\theta\right]^{a + 2\pi}_a = \begin{cases} 
      \theta - \floor*{\frac{\theta - a}{2\pi}} \cdot 2\pi & \quad \text{if } \theta \geq 0, \\
      \theta + \left(\floor*{\frac{-\theta -a}{2\pi}} + 1 \right)\cdot 2\pi
      & \quad \text{if } \theta < 0, \\
   \end{cases}
\]
where the floor operator, which returns the largest integer smaller than the argument, is denoted by $\floor*{\cdot}$ \citep{benjamin2017autonomous}.

Second, $\alpha_{OS}^{TS}$, the relative bearing from the OS to the TS, is computed as the difference between the absolute bearing, $\beta_{OS}^{TS}$, and the OS's heading: $\alpha_{OS}^{TS} = \left[\beta_{OS}^{TS} - \psi_{OS}\right]^{2\pi}_0$. \rd{Conversely}, $\alpha_{TS}^{OS}$ and $\beta_{TS}^{OS}$ are the quantities from the perspective of the TS toward the OS. \rd{Scenarios can be classified based on} the heading intersection angle and the relative bearing, \rd{as outlined in} Table \ref{tab:COLREG_encounter_definitions}, and \rd{they are visualised} in Figure \ref{fig:COLREG_viz}. Note that the rules should always be seen in \rd{the} context \rd{of} the present situation at sea, and seafarers consider softly defined practices referred to as good seamanship.

Moreover, the definitions above focus solely on situations with two vessels. In practice, multi-ship situations are possible, especially in highly congested sea areas. A reliable control system needs to effectively process \rd{the} information of all relevant target ships to find a safe path in \rd{these} cases, despite the insufficiently defined rule set \citep{kang2021collision}.

\def\arraystretch{2.2}
\begin{table}
\setlength{\tabcolsep}{6pt}
\footnotesize
    \centering
    \begin{tabular}{llcl}
         Situation & $\sigma$ & Requirements & Action from OS \\
         \toprule
         \rd{Head-on}  & 1 & \makecell[l]{\hspace{0.12cm} $\alpha_{OS}^{TS} \in \{[0^{\circ}, 5^{\circ}] \cup [355^{\circ}, 360^{\circ})\}$ \\ \hspace{0.3cm}  $C_T \in [175^{\circ}, 185^{\circ}]$} & \makecell[l]{Alter course to starboard,\\ pass TS on its portside}\\
         \midrule
         \makecell[l]{Starboard\\ crossing} & 2 & \makecell[l]{\hspace{0.12cm} $\alpha_{OS}^{TS} \in [5^{\circ}, 112.5^{\circ}]$ \\\hspace{0.12cm}  \hspace{0.07cm} $C_T \in [185^{\circ}, 292.5^{\circ}]$} & \makecell[l]{Alter course to starboard,\\ avoid crossing ahead of TS}\\
         \midrule
         Port crossing & 3 & \makecell[l]{\hspace{0.12cm}  $\alpha_{OS}^{TS} \in [247.5^{\circ}, 355^{\circ}]$ \\ \hspace{0.3cm}  $C_T \in [67.5^{\circ}, 175^{\circ}]$} & \makecell[l]{Keep course,\\ TS is give-way vessel}\\
         \midrule
         Overtaking & 4 & \makecell[l]{\hspace{0.12cm}  $\alpha_{TS}^{OS} \in [112.5^{\circ}, 247.5^{\circ}]$ \\ \hspace{0.07cm} \hspace{0.12cm} $C_T \in \{[0^{\circ}, 67.5^{\circ}] \cup [292.5^{\circ}, 360^{\circ})\}$ \\ $U_{OS,R} > U_{TS}$} & \makecell[l]{Overtake \rd{on} any side,\\ keep out of TS's way}\\
    \end{tabular}
    \caption{Classification of COLREG encounter situations between two vessels following \cite{xu2020intelligent}. We define the variable $\sigma$ to refer to a particular scenario. If no requirements \rd{are} fulfilled, we set $\sigma = 0$. Overtaking imposes\rd{, in addition to} $C_T$ and $\alpha_{OS}^{TS}$\rd{,} a third constraint, namely that the relative speed of the OS in the direction of the TS's course, $U_{OS,R}$, is larger than the TS's total speed $U_{TS}$.}
    \label{tab:COLREG_encounter_definitions}
\end{table}

\subsection{Collision risk assessment}
\subsubsection{Ship domain and CPA}\label{subsubsec:approaches_CR}
Assessing the collision risk with other vessels is required by the COLREGs and constitutes a fundamental part of an ASV. The methodological repertoire of the literature is diverse and has recently been reviewed by \cite{ozturk2019individual} and \cite{huang2020ship}. Two crucial approaches are the definition of ship domains \citep{goodwin1975statistical, smierzchalski2005ships, szlapczynski2017review} and the concept of \rd{the} closest point of approach \citep{mou2010study, zhao2019colregs}. The ship domain is a safe area around the vessel, which should not be entered by other ships. Several geometric representations of ship domains have been considered, and they are primarily asymmetric\rd{,} with \rd{a} larger space on the starboard side to enable COLREG-compliant \rd{COLAV}. While there are several possibilities for defining a collision based on \rd{the} ship domain \citep{heiberg2022risk}, in this study, we define the TS's midship position being at or inside the OS's ship domain as a collision event. While \rd{the} non-violation of the ship domain equates to reliable \rd{COLAV}, the approach lacks a quantitative \rd{collision risk} metric since \rd{the} violation or non-violation of the domain is a binary variable \citep{ha2021quantitative}. 

\begin{figure}
    \centering
    \begin{subfigure}[c]{0.3\textwidth}
    \vspace{40pt}
        \includegraphics[width=0.8\linewidth]{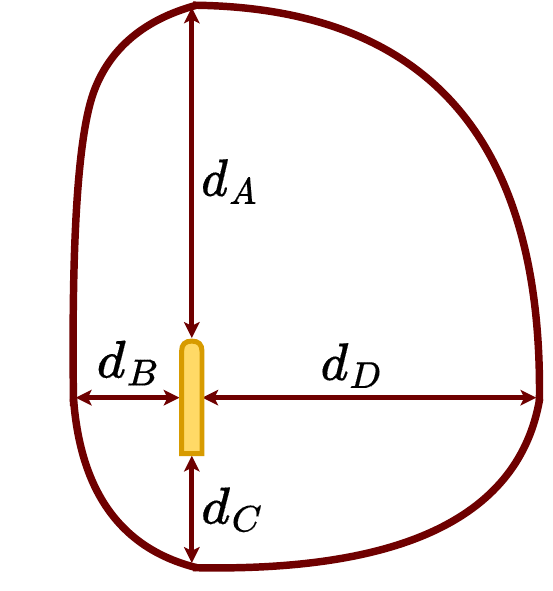}
    \end{subfigure}
    \begin{subfigure}[c]{0.45\textwidth}
        \includegraphics[width=\linewidth]{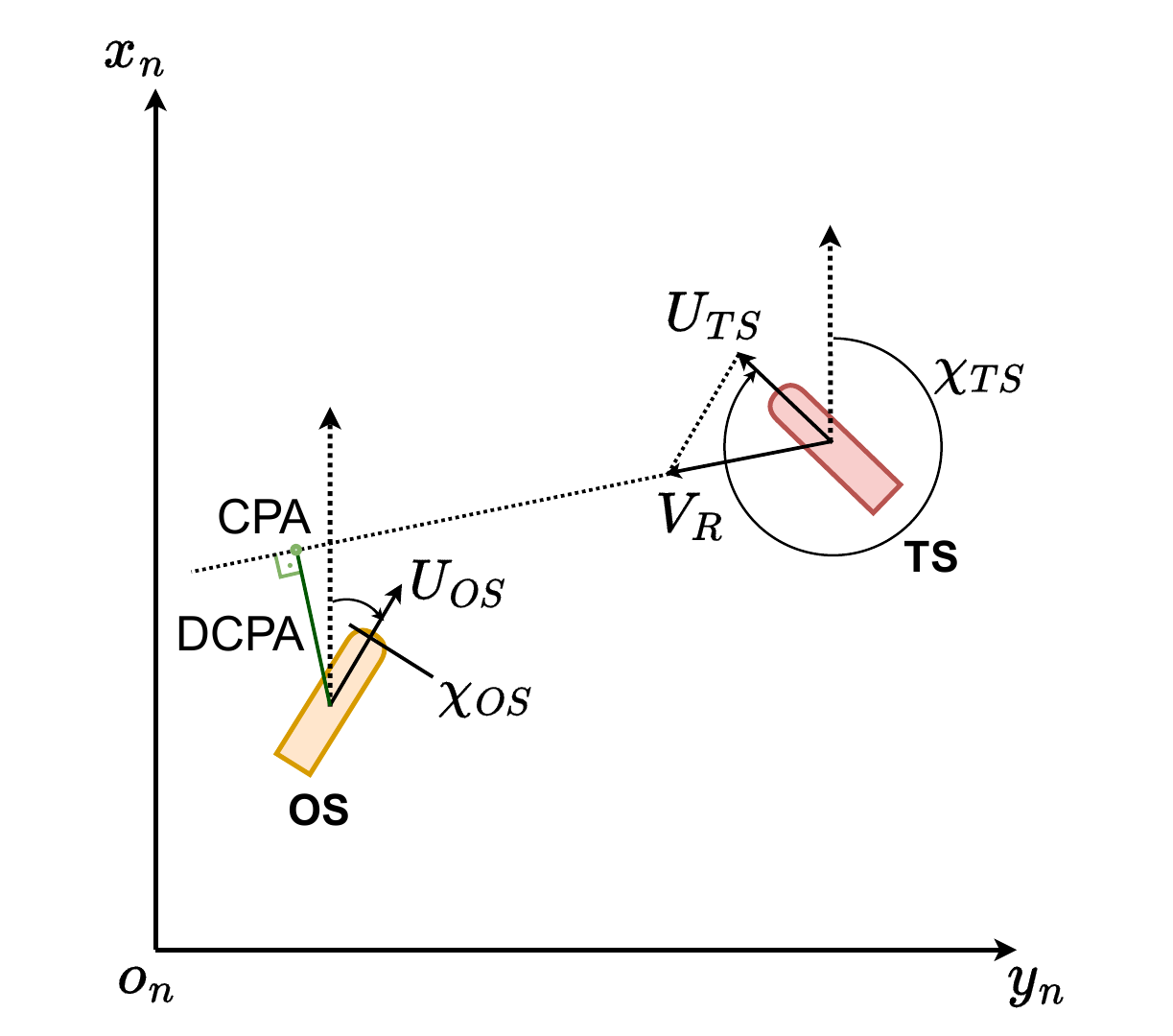}
    \end{subfigure}
    \caption{\rd{Visualisation} of the collision risk assessment methods used in this work. The left-hand side shows the asymmetric ship domain of \cite{chun2021deep}, which we construct as a combination of four \rd{ellipses} with $d_A = d_D = 3 L_{pp}$ and $d_B = d_C = L_{pp}$. The right-hand side \rd{visualises} the \rd{concepts of the} CPA and DCPA based on \cite{heiberg2022risk}. In this illustration, the heading equals the course angle for both vessels.}
    \label{fig:domain_CPA}
\end{figure}

Filling this gap, the CPA concept describes the closest point two vessels will encounter under the assumption that both keep their course and speed. Two key quantities are the time until the CPA \rd{is reached}, called \rd{the} TCPA, and the distance between the two vessels at the CPA, called \rd{the} DCPA \citep{lenart1983collision}:

\begin{equation}\label{eq:DCPA_TCPA_lenart}
\begin{aligned}
    \text{TCPA} &= -\frac{(x_{n,TS} - x_{n,OS}) \cdot v_{r, x_{n}} + (y_{n,TS} - y_{n,OS}) \cdot v_{r, y_{n}}}{v_{r, x_{n}}^2 + v_{r, y_{n}}^2},\\
    \Delta x_{n,\text{TCPA}} &= (x_{n,OS} + \text{TCPA} \cdot v_{x_{n,OS}}) - (x_{n,TS} + \text{TCPA} \cdot v_{x_{n,TS}}),\\
    \Delta y_{n,\text{TCPA}} &= (y_{n,OS} + \text{TCPA} \cdot v_{y_{n,OS}}) - (y_{n,TS} + \text{TCPA} \cdot v_{y_{n,TS}}),\\
    \text{DCPA} &= \sqrt{ (\Delta x_{n,\text{TCPA}})^2 + (\Delta y_{n,\text{TCPA}})^2},
\end{aligned}
\end{equation}
where $x_{n,OS}$, $y_{n,OS}$, $v_{x_{n,OS}}$, \rd{and} $v_{y_{n,OS}}$ are the positions and speeds of the OS in $\{n\}$, respectively. The notation for the TS is \rd{similar}, and the relative speeds are $v_{r, x_{n}} = v_{x_{n,TS}} - v_{x_{n,OS}}$ and $v_{r, y_{n}} = v_{y_{n,TS}} - v_{y_{n,OS}}$. The drawback of the CPA-based metrics is their inability to reliably guarantee \rd{COLAV} when \rd{analysed} in isolation \citep{ha2021quantitative} and the abstraction from present yaw moments and future non-linear \rd{behaviour}. Figure \ref{fig:domain_CPA} illustrates \rd{these} concepts.

\subsubsection{Construction of a collision risk metric}\label{subsubsec:CR_metric}
The construction of a reliable collision risk metric is of central importance for an RL agent since it directly signals the criticality of an opposing ship. Based on the concepts from \rd{subsection} \ref{subsubsec:approaches_CR}, we develop a novel $CR$ metric, unifying the concepts $CR_{\rm CPA}$ and $CR_{\rm ED}$\rd{, which will} be defined later, as follows:
\begin{equation}\label{eq:CR_metric}
CR = 
\begin{cases}
      1 & \quad \text{if TS in ship domain of OS}, \\
      \max(CR_{\rm CPA}, CR_{\rm ED})
      & \quad \text{otherwise.} \\
\end{cases}
\end{equation}

\rd{For all situations, $CR \in [0,1]$,} where 0 suggests no risk of collision and 1 is a collision event. Through the maximum operator, we consider the more critical of \rd{the two} components $CR_{\rm CPA}$ and $CR_{\rm ED}$. The first \rd{component}, $CR_{\rm CPA}$, signals a large collision risk if \rd{the} TCPA (absolute value) and DCPA are small since the OS then faces a potentially dangerous situation in the near future. The second component, $CR_{\rm ED}$, \rd{considers that} a small Euclidean distance between two ships \rd{indicates} a high risk of collision. The necessity of introducing this second component is discussed in detail below. 

\begin{figure}
    \centering 
    \includegraphics[width=0.9\textwidth]{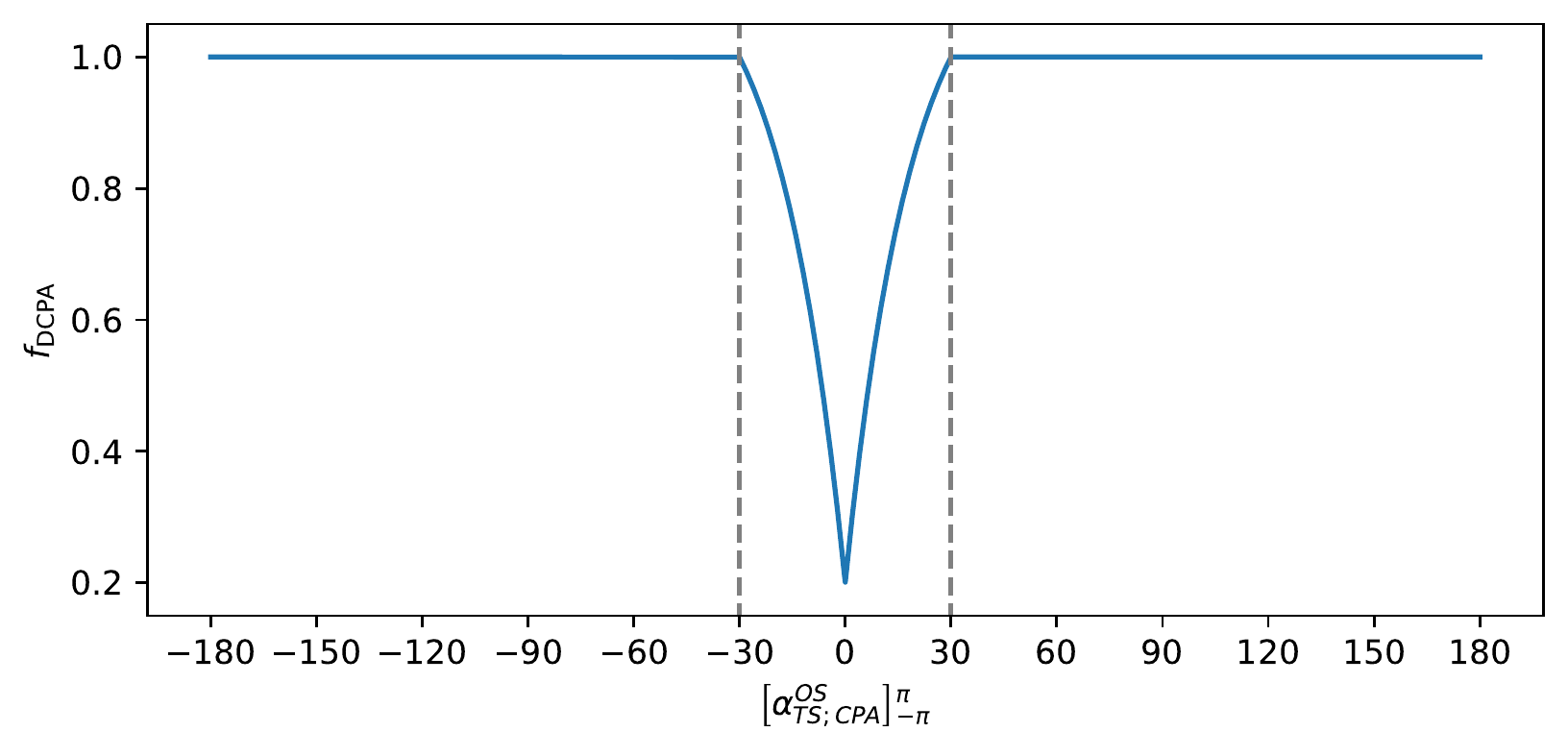}
    \caption{Specification of the factor $f_{\rm DCPA}$ for bow-crossing avoidance. The illustration shows the case \rd{in which} $\rm TCPA \geq 0$, and $[\alpha^{\rm OS}_{\rm TS;CPA}]^{\pi}_{-\pi}$ \rd{is given} in \rd{degrees}.}
    \label{fig:f_DCPA}
\end{figure}

Building on \cite{mou2010study}, we define $CR_{\rm CPA}$ to behave exponentially in negative TCPA and DCPA:
\begin{equation}\label{eq:cr_cpa}
    CR_{\rm CPA} = \exp\left\{c_1 \cdot \left[ \text{DCPA}' + c_{2}^{\mathbf{1}\{\text{TCPA} \geq 0\}} \cdot c_{3}^{\mathbf{1}\{\text{TCPA} < 0\}} \vert \text{TCPA} \vert \right]\right\},
\end{equation}
where we set $c_1 = \log(0.1) / 3704$, $c_2 = 1.5$\rd{, and} $c_3 = 20$ in our simulation. Note that \rd{\unit[3704]{m} is equal to} 2 nautical miles (NM) and that the sign of $c_1$ is negative. The indicator function $\mathbf{1}\{x\}$ is 1 if $x$ is true and 0 otherwise. Since we set $c_2 < c_3$, we \rd{weight} positive and negative TCPA \rd{values} differently and achieve a rapid decay in \rd{the} collision risk after the CPA has been passed while avoiding a jump in the metric by, e.g. setting the $CR_{\rm CPA}$ component \rd{to} zero. Furthermore, instead of directly using the DCPA \rd{value} from (\ref{eq:DCPA_TCPA_lenart}), we consider the distance of the TS to the OS's ship domain via the following modification: 
\begin{equation}\label{eq:DCPA_prime}
    \text{DCPA}' = f_{\rm DCPA} \cdot \max\left[0,\text{DCPA} - D(\alpha^{\rm TS}_{\rm OS, CPA})\right].
\end{equation}
The function \rd{$D: [0, 2\pi) \rightarrow \mathbb{R}$ returns the distance to the} ship domain for a given encounter angle, which is non-constant since we consider an asymmetric shape \rd{(}see Figure \ref{fig:domain_CPA}\rd{)}. \rd{Furthermore}, we denote the relative bearing from the perspective of the OS towards the TS at the CPA, or equivalently, at $\rm TCPA = 0$, with $\alpha^{\rm TS}_{\rm OS, CPA}$. Thus, we consider the safety area of the OS at the CPA. \rd{Moreover}, we want to explicitly avoid \rd{the scenario in which} the agent crosses at the bow of other ships, which is considered bad practice at sea. The factor $f_{\rm DCPA}$ is responsible for \rd{this}:
\[ f_{\rm DCPA} = \begin{cases} 
      c_4 - \exp(c_5 \cdot \vert [\alpha^{\rm OS}_{\rm TS;CPA}]^{\pi}_{-\pi}\vert) & \quad \text{if $\rm TCPA \geq 0$ and $\vert [\alpha^{\rm OS}_{\rm TS;CPA}]^{\pi}_{-\pi}\vert \leq \frac{\pi}{6}$}, \\
      1 & \quad \text{otherwise,} \\
   \end{cases}
\]
where $\alpha^{\rm OS}_{\rm TS, CPA}$ is the relative bearing from the perspective of the TS towards the OS at the CPA, and $c_4 = 1.2$ \rd{and} $c_5 = -\log(5)/\frac{\pi}{6}$ are constants. \rd{This} factor \rd{penalises} the undesired crossing \rd{behaviour} by decreasing the sensed distance at the CPA and thus increasing the risk of collision. Figure \ref{fig:f_DCPA} \rd{visualises} the \rd{behaviour} of \rd{this} factor.

The component $CR_{\rm CPA}$ is a reasonable quantification of \rd{the} collision risk in the majority of scenarios. However, there are distinct disadvantages that made the inclusion of the component $CR_{\rm ED}$ based on the Euclidean distance between the ships, denoted $d_{\text{OS}}^{\text{TS}}$, necessary. Consider a scenario \rd{in which} two ships are close but have almost the same course. The CPA-based metrics signal that this situation is not risky since the CPA either \rd{lies} a long time in the past or is far in the future. However, if a ship turns even slightly, the absolute value of the TCPA jumps to a much smaller value, and suddenly the situation is precarious. To prevent this drawback, we define the second component as follows:
\begin{equation}\label{eq:CR_ED}
    CR_{\rm ED} = \exp\left\{c_{6}^{\rd{-1}} \cdot \left[d_{\text{OS}}^{\text{TS}} - D(\alpha_{\rm OS}^{\rm TS})\right] \right\},
\end{equation}
where $c_6 = -0.3 \cdot 3704 = -1111.2$ is a negative constant. The subtraction of the \rd{distance to the} ship domain is motivated as in (\ref{eq:DCPA_prime}). 

We \rd{emphasise} that the functional relationships (\ref{eq:CR_metric})--(\ref{eq:CR_ED}) build a robust collision risk metric for vessels of various classes. However, we set the constants $c_1, \ldots, c_6$ specifically for the KVLCC2 tanker, and these values should be adjusted for a vessel with different characteristics and manoeuvrability. Moreover, similar to how human seafarers have different interpretations of safe passing distances \citep{miyoshi2022rules}, the ASV designer can calibrate $c_1, \ldots, c_6$ depending on how \rd{risk-averse} the agent should be.

\sloppy

\section{Reinforcement learning methodology}\label{sec:RL_methods}
\subsection{Background}\label{subsec:RL_background}
Reinforcement learning is a methodological ensemble in which an agent learns based on trial and error in an environment \citep{sutton2018reinforcement}. The common formalism of the problem is a Markov \rd{decision process} (MDP; \citealp{puterman1994markov})\rd{, which is represented by} the tuple $(\mathcal{S}, \mathcal{A}, \mathcal{P}, \mathcal{R}, \gamma)$. $\mathcal{S}$ is the state space, $\mathcal{A}$ is the discrete action space, $\mathcal{P}: \mathcal{S} \times \mathcal{A} \times \mathcal{S} \rightarrow [0,1]$ is the state transition probability distribution, $\mathcal{R}: \mathcal{S} \times \mathcal{A} \rightarrow \mathbb{R}$ is a bounded reward function, and $\gamma \in [0,1)$ is the discount factor. At each step $t$, the agent receives state information $s_t \in \mathcal{S}$, takes an action $a_t \in \mathcal{A}$ according to a policy $\pi: \mathcal{S} \times \mathcal{A} \rightarrow [0,1]$, gets a reward $r_t$ generated by $\mathcal{R}$, and transitions according to $\mathcal{P}$ to the next state $s_{t+1} \in \mathcal{S}$. Transferred to our maritime application case, the state includes the positional and motion information of the ships, while the action is the change \rd{in} the rudder angle \rd{of} the OS.

The discounted sum of rewards during an episode is called \rd{the} return. Many frequently used RL algorithms define action-values: $Q^{\pi}(s,a) = \operatorname{E}\left[\sum_{t=0}^{\infty} \gamma^t r_t \vert s_0 = s, a_0 = a \right]$. Thus, these so-called $Q$-values describe the expected return when \rd{action $a$ in state $s$ is executed and policy $\pi$ is followed afterward}. The methodological basis of our work is the \rd{$Q$-learning} algorithm of \cite{watkins1992q}. The objective of the algorithm is to attain the optimal action-values $Q^*(s,a)$ for all $s \in \mathcal{S}, a \in \mathcal{A}$, which yield an optimal policy $\pi^{*}(s) = \argmax_{a' \in \mathcal{A}} Q^*(s,a)$ for all $s \in \mathcal{S}, a \in \mathcal{A}$ if such an optimal policy exists; see \cite{puterman1994markov} for a deep discussion. The convergence conditions for \rd{$Q$-learning} are relatively mild \citep{tsitsiklis1994asynchronous}.

Storing a $Q$-value for each state-action pair is infeasible for continuous state spaces, which occur in our maritime environment. \cite{mnih2015human} proposed the DQN algorithm, an extension of \rd{$Q$-learning in which} action-values are approximated by deep neural networks. More precisely, we consider the function $Q(s,a;\theta)$ for all $s \in \mathcal{S}, a \in \mathcal{A}$, where $\theta$ is the parameter set of the \rd{neural network}. The \rd{optimisation} is performed \rd{using} gradient \rd{descent}:
\begin{equation*}
    \theta \leftarrow \theta + \tau \left[y -Q(s,a;\theta) \right] \nabla_{\theta} Q(s,a;\theta), 
\end{equation*}
where $y = r + \gamma \max_{a' \in \mathcal{A}} Q(s',a';\theta^{-})$. The successor state after action $a$ \rd{is executed} in state $s$ is denoted $s'$, $\tau$ is the learning rate, and $\theta^{-}$ is the parameter set of the target network, a time-delayed copy of $\theta$. Moreover, the DQN uses experience replay to \rd{stabilise the} training.

\subsection{Spatial-temporal recurrent architecture}\label{subsec:spat_temp_arch}
In practice, observing the full state of a system is often not possible due to sensor limitations, noise, time delays, and other factors. Acknowledging this circumstance, we formally consider a \rd{partially observable} MDP (POMDP; \citealp{kaelbling1998planning}), which extends the MDP-tuple $(\mathcal{S}, \mathcal{A}, \mathcal{P}, \mathcal{R}, \gamma)$ of \rd{subsection} \ref{subsec:RL_background} by \rd{adding} two components: the observation space $\mathcal{O}$ and the observation function $\mathcal{Z}: \mathcal{S}\times \mathcal{A}\times \mathcal{O}\rightarrow [0,1]$. At time $t$, instead of receiving the full state $s_t \in \mathcal{S}$, the agent \rd{receives} an observation $o_t \in \mathcal{O}$, which is generated by $\mathcal{Z}$. On an implementation level\rd{,} for the DQN, the action-value function \rd{receives} as input an observation $o \in \mathcal{O}$ instead of the state $s \in \mathcal{S}$.

Recently, \cite{meng2021memory} proposed the LSTM-TD3 algorithm, an extension of the TD3 algorithm of \cite{fujimoto2018addressing}\rd{,} to deal with POMDPs. A memory extraction component based on a long short-term memory (LSTM; \citealp{hochreiter1997long}) architecture processes several past observations, and \rd{it} combines \rd{this} information with the observation of the current time step. Thus, the LSTM-TD3 incorporates a \emph{temporal recurrency}, which \rd{allows} information \rd{from} observation sequences over time \rd{to be} processed. 

In our maritime application case, an observation $o_t = o_{OS,t} \cup o_{TS,t}$ consists of two components: $o_{OS,t}$, the information regarding the navigational status of the OS, and $o_{TS,t}$, the information of \rd{the} surrounding target ships. The observation $o_{TS,t}$ consists of features inspired by AIS data and will be \rd{described in detail in subsection} \ref{subsec:observations_actions}. Importantly, the size of $o_{TS,t}$ is $N_t \cdot 6$, where $N_t$ is the number of the target ships surrounding the OS at time $t$. \rd{Crucially, $N_t$ can change over time; it cannot be processed with a standard fully connected neural network that requires a fixed input size.} Building on \rd{the work of} \cite{everett2018motion, everett2021collision}, we propose a more flexible approach by introducing a \emph{spatial recurrence} that sequentially processes the target ships according to their estimated collision risk. Similar to the temporal LSTM\rd{, which loops} over sequences of observations of different time steps, the introduced spatial LSTM loops over the sub-observations of different target ships \emph{at the same point \rd{in} time}.

\begin{figure}[ht]
    \centering
    \includegraphics[width=0.8\textwidth]{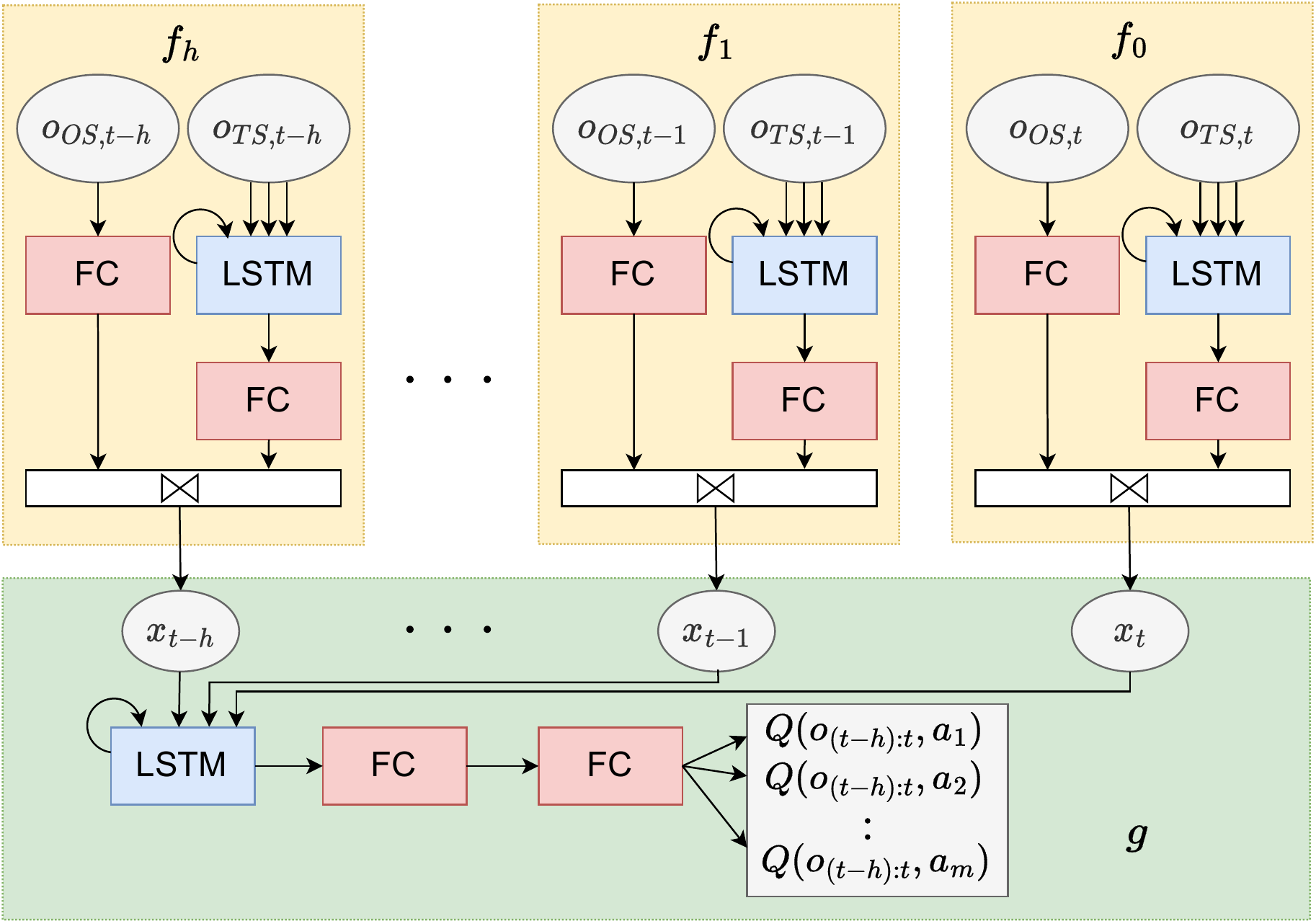}
    \caption{Spatial-temporal recurrent architecture. The concatenation symbol is $\bowtie$, and FC refers to a fully connected layer with 64 output features\rd{,} as specified in deep learning frameworks like PyTorch \citep{paszke2019pytorch}. Thus, strictly speaking, it is one weight matrix rather than a layer. We use ReLU activations after each FC block except for the last one, which applies the identity function to yield $Q$-value estimates in $\mathbb{R}^M$. The number of hidden units in the LSTM blocks is 64.}
    \label{fig:Dual_LSTM_Architecture}
\end{figure}

In the proposed architecture, we implement both \emph{temporal} and \emph{spatial recurrency} components. Thus, we can deal with an arbitrary number of target ships while efficiently processing information over several time steps. Formally\rd{,}
\begin{align*}
    x_{t-l} &= f_l(o_{OS,t-l}, o_{TS,t-l}; \theta_{f_l}) \quad \text{for} \quad l = 0, \ldots, h,\\
    Q(o_{(t-h):t}, a_i; \theta) &= g(x_{t-h}, \ldots, x_{t-1}, x_t, a_i; \theta_g) \quad \text{for} \quad i = 1, \ldots, M,
\end{align*}
where $M$ is the cardinality of the action space and $o_{(t-h):t} = \cup_{l=0}^{h} o_{t-l}$. The number of considered past observations is $h$, and we fix $h = 2$ throughout the paper since it provides a very good performance. The functions $f_l$ with parameters $\theta_{f_l}$ for $l = 0,\ldots,h$ represent the spatial recurrency, and the function $g$, \rd{parametrised} by $\theta_g$, corresponds to the temporal recurrency. The complete parameter set is $\theta = \left(\cup_{l=0}^{h} \theta_{f_l}\right)\cup \theta_g$ and the architecture is illustrated in Figure \ref{fig:Dual_LSTM_Architecture}. The algorithmic procedure is identical to the DQN outlined in \cite{mnih2015human}, and the approach is compatible with further modifications like \rd{those of} \cite{schaul2015prioritized}, \cite{van2016deep}, or \cite{waltz2022two}. \rd{It is worth noting that there are different ways to incorporate temporal recurrence into the architecture. While our method takes advantage of the promising outcomes demonstrated by \cite{meng2021memory} by processing the current observation alongside the previous $h$ time steps, other studies directly store and process the entire history of the episode in recurrent layers \citep{heess2015memory, wang2019autonomous}. Although this represents an interesting area for future research, our current approach delivers a robust performance, and we chose to maintain it for the scope of this paper.}

\subsection{Observation and action spaces}\label{subsec:observations_actions}
We define $o_{OS,t}$, the OS-related component of the observation at time $t$, by considering the OS's velocity information \rd{and} rudder angle, together with the Euclidean distance $d_{\text{OS},t}^{\text{G}}$ and relative bearing $\alpha^{\rm G}_{\rm OS,t}$ to the goal $(x_{n, G}, y_{n, G})$:
\begin{equation*}
       o_{OS,t} = \left(\frac{u_t}{u_{\rm scale}}, \frac{v_t}{v_{\rm scale}}, \frac{\Tilde{r}_t}{\Tilde{r}_{\rm scale}}, \frac{\dot{\Tilde{r}}_t}{\dot{\Tilde{r}}_{\rm scale}}, \frac{\delta_t}{\delta_{\rm scale}}, \frac{d_{\text{OS},t}^{\text{G}}}{d_{\rm scale}}, \frac{[\alpha^{\rm G}_{\rm OS,t}]^{\pi}_{-\pi}}{\pi} \right)^{\top}.
\end{equation*}
The denominators are scaling factors: $u_{\text{scale}} = \unit[7]{m/s}$, $v_{\text{scale}} = \unit[0.7]{m/s}$, $\Tilde{r}_{\text{scale}} = \unit[0.004]{rad/s}$, $\dot{\Tilde{r}}_{\text{scale}} = \unit[8\cdot 10^{-5}]{rad/s^2}$, \rd{$\delta_{\text{scale}} = 20{^\circ}$, and $d_{\text{scale}} = \unit[14]{NM}$} to achieve inputs approximately in the interval $[-1,1]$. To define $o_{TS,t}$, the feature vector \rd{of} the target ships at time $t$, we assume \rd{that we have} access to \rd{the} AIS data of \rd{other} ships to \rd{obtain} their course, speed, and positional information, which serve as \rd{the} input to our \rd{deep $Q$-network}. Precisely, the component is defined as \rd{follows}:
\begin{equation*}
    o_{TS,t} = \left(
    o_{1,t}, \ldots, o_{N_t, t}
    \right)^{\top},
\end{equation*}
where $o_{i,t}$\rd{,} with $i=1,\ldots,N_t$\rd{,} is the information about the $i$th target ship at time $t$\rd{; the $o_{i,t}$ vectors} are sorted with respect to \emph{ascending collision risk}, following the newly defined metric in \rd{subsection} \ref{subsubsec:CR_metric}. \rd{Again, the} size of $o_{TS,t}$ depends on $N_t$. The feature vector for TS $i$ is defined as
\begin{equation*}
    o_{i,t} = \left(\frac{[C_{T,i,t}]^{\pi}_{-\pi}}{\pi}, \frac{U_{i,t}}{u_{\rm scale}}, \frac{d_{\text{OS},t}^{\text{i}} - D(\alpha_{\rm OS,t}^{\rm i})}{d_{\rm scale}}, \frac{[\alpha^{i}_{OS,t}]^{\pi}_{-\pi}}{\pi}, \sigma_{i,t}, CR_{i,t} \right)^{\top}\rd{,}
\end{equation*}
where $\sigma_{i,t}$ is the COLREG-encounter situation from Table \ref{tab:COLREG_encounter_definitions}. The element $CR_{i,t}$ is our collision risk metric for ship $i$ at time $t$. Crucially, if no target ship is present, our recurrent structure still requires information about at least one ship. In this case, we artificially create a no-risk target ship $o_{TS,t} = o_{1,t} = (-1,0,1,-1,0,0)^{\top}$. This technical padding procedure is unavoidable when dealing with neural networks. However, due to our recursive structure, we need to pad information about one ship maximally instead of padding the whole surrounding area, as is the case when specifying a non-recursive architecture.

Regarding the action space, the agent can only control changes in the rudder angle by selecting one of three actions: $a_t \in \{0, -\Delta \delta, \Delta \delta\}$. We set $\Delta \delta = 5^{\circ}$ in our simulation, which builds\rd{,} together with our simulation step size of \unit[3]{\rd{s},} a realistic steering \rd{behaviour} of $\rd{\sim}$$1.67 ^{\circ}/\rm s$. \rd{Furthermore}, we clip the rudder angle to an absolute value of $20^{\circ}$. We do not allow thrust control to keep the action space as simple as possible, and, in practice, steering is generally preferred over thrust changes, especially for a large tanker like the KVLCC2. Thus, we keep the revolutions per second of the propeller fixed to 1.8 for the OS, which results in a speed of $\rd{\sim}$$\unit[7.42]{m/s}$ without steering.

\subsection{Reward design}
\rd{The reward function plays a crucial role in RL applications, as it is used to obtain the desired behaviour of the agent. While we aimed to keep the reward function as simple as possible, we identified five key components that were necessary for developing a robust end-to-end agent that could generate appropriate rudder angles from AIS data. These components are a heading and distance reward towards the goal, a collision penalty, a traffic rule component, and a comfort reward.} We must admit that many parameters were experimentally obtained in order to obtain the best performance. 

Building on \rd{the work of} \cite{xu2022path}, the heading and distance reward \rd{enables} the agent to construct a path towards the goal:
\begin{equation*}
    r_{\text{dist},t} = \frac{d_{\text{OS},t-1}^{\text{G}} - d_{\text{OS},t}^{\text{G}}}{c_7} + c_8, \qquad \quad r_{\text{head}, t} = -  \frac{\vert[\alpha^{\rm G}_{\rm OS,t}]^{\pi}_{-\pi}\vert}{\pi},
\end{equation*}
with constants $c_7 = 20$ \rd{and} $c_8 = -1$. \rd{Note that the speed of the OS is constant at $\sim$$\unit[7.42]{m/s}$ and that our simulation step size is $\unit[3]{s}$. Thus, if the OS is moving straight toward the goal, we have $d_{\text{OS},t-1}^{\text{G}} - d_{\text{OS},t}^{\text{G}} \approx \unit[3]{s} \cdot \unit[7.42]{m/s} = \unit[22.26]{m}$. Consequently, with the given values of $c_7$ and $c_8$, the distance reward is in $[-2.113, 0.113]$, while the heading reward is in $[-1, 0]$. Such normalisations make it easier to interpret the agent's final performance and stabilise the training process.} The third reward constituent \rd{penalises the} collision risk by considering all other vessels via $r_{\text{coll},t} = \sum_{i=1}^{N_t} r_{\text{coll},i,t}$, where the component for TS $i$ is
\begin{equation*}
    r_{\text{coll},i,t} = 
    \begin{cases}
      c_9 & \quad \text{if } CR_{i,t} = 1,   \\  
      -\sqrt{CR_{i,t}} & \quad \text{otherwise,} \\
    \end{cases}
\end{equation*}
where $c_9 = -10$. Using the square root in the reward calculation deviates from prior linear suggestions, e.g. \cite{chun2021deep}. We found this adjustment useful because, since $CR_{i,t} \in [0,1]$, it \rd{penalises} the risk of a collision earlier and \rd{incentivises} the agent to develop a more foresighted \rd{behaviour}. \rd{Furthermore}, we include a large negative penalty in \rd{the} case of a collision. 

The fourth component is responsible for COLREG compliance since it \rd{penalises} the agent if it does not turn right in \rd{the} head-on and starboard crossing situations for vessels \rd{for which} the CPA has not been passed yet. We set $r_{\text{COLREG},t} = \sum_{i=1}^{N_t} r_{\text{COLREG},i,t}$, with the component for TS $i$ being

\begin{equation*}
    r_{\text{COLREG},i,t} = 
    \begin{cases}
      c_{10} & \quad \text{if } \text{TCPA}_{i,t} \geq 0 \text{ and } \Tilde{r}_t < 0 \text{ and } \sigma_{i,t} \in \{1,2\},   \\  
      0 & \quad \text{otherwise,} \\
    \end{cases}
\end{equation*}
for the constant $c_{10} = -1$. We chose the yaw rate as a criterion instead of the rudder angle (as in \cite{xu2020intelligent}) since the yaw rate is, by definition, the change in \rd{the} heading and quantifies the actual turning of the ship's nose. In ship dynamics, there can be a delay of several time steps in the translation from a sign change in \rd{the} rudder angle to a sign change in \rd{the} yaw rate, making the rudder angle less \rd{useful when it comes to judging} whether a ship turns correctly.

The final reward quantity is an adaptive comfort reward that \rd{penalises} steering if there is no or only a small risk of collision. Precisely, we have

\begin{equation}\label{eq:r_comf}
    r_{\text{comf},t} = 
    \begin{cases}
        c_{11} & \quad \text{if } a_t \neq 0 \text{ and } \forall i \in 1,\ldots,N_t: CR_{i,t} \leq c_{12},\\
        0 & \quad \text{otherwise,}
    \end{cases}
\end{equation}
where we set $c_{11} = -1$ and $c_{12} = 0.2$ in our simulation. We want to achieve a moderate and practically possible steering \rd{behaviour} with the comfort reward, which avoids a non\rd{-}human succession of \rd{very} frequent course alterations. The total reward at a step $t$ is constructed as \rd{follows}:
\begin{equation*}
    r_t = r_{\text{dist},t}w_{\text{dist}} + r_{\text{head},t} w_{\text{head}} + r_{\text{coll},t}w_{\text{coll}} + r_{\text{COLREG},t}w_{\text{COLREG}}+ r_{\text{comf},t} w_{\text{comf}},
\end{equation*}
for weights $w_{\text{dist}} = \frac{0.05}{6.15} \approx 0.0081, w_{\text{head}} = \frac{2.0}{6.15} \approx 0.3252, w_{\text{coll}} = \frac{1.8}{6.15}\approx 0.2927, w_{\text{COLREG}} = \frac{2.0}{6.15}\approx 0.3252$, and $w_{\text{comf}} = \frac{0.3}{6.15}\approx 0.0488$. The weights have been identified experimentally by running a grid search over different configurations.

\section{Training}\label{sec:training}
\subsection{Environmental setup}\label{subsec:env_setup}
\begin{figure}[h]
    \centering
    \includegraphics[width=\textwidth]{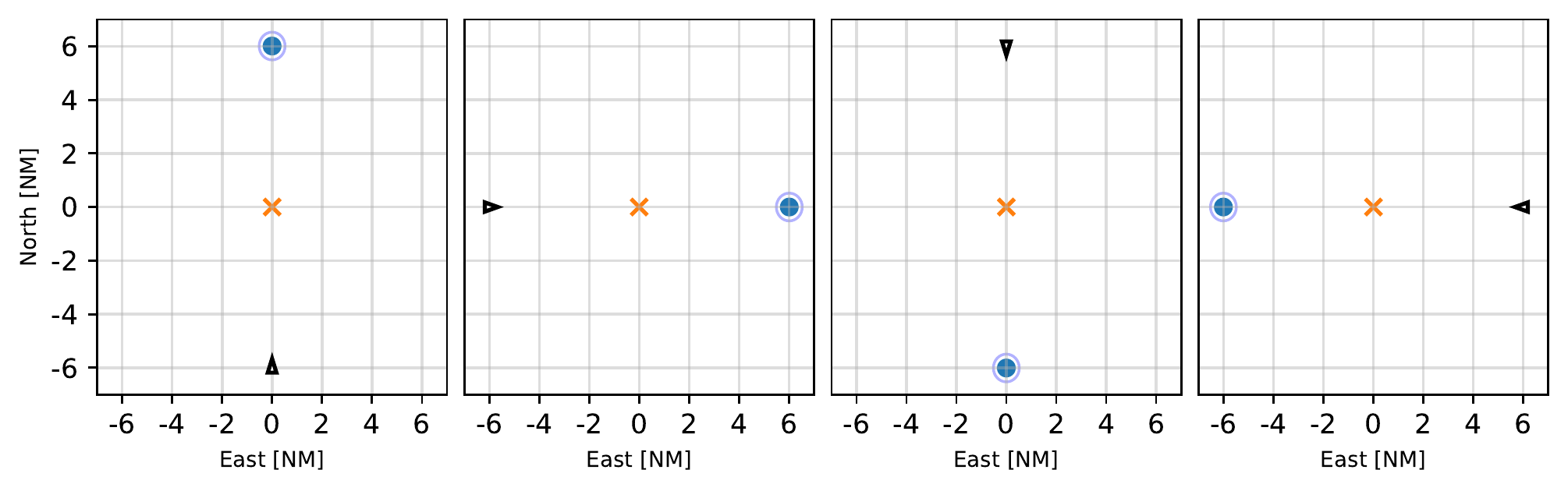}
    \caption{OS and goal spawning constellations. The black triangle is the OS, the orange cross is $o_n$, and the blue dot is the goal. The blue circle around the goal indicates the area (with radius $3 L_{pp}$) \rd{that} we consider as reaching the goal.} 
    \label{fig:goal_positions}
\end{figure}

We consider a simulation environment with $x_n, y_n \in [\unit[-7]{NM}, \unit[7]{NM}]$. First, we uniformly sample an OS heading from $\{0, \frac{\pi}{2}, \pi, \frac{3}{2} \pi\}$. \rd{Then,} the position of the OS in $\{n\}$ is set so that \rd{after} \unit[25]{\rd{minutes}} in simulation time, \rd{it is} at $o_n=(0,0)$ if no steering takes place. The goal coordinate is initiated at the same distance but mirrored on $o_n$. Once \rd{the} OS and goal are set, to improve \rd{generalisation}, we randomly disturb the OS's heading \rd{using} a \rd{realisation} of a uniform distribution $\mathcal{U}(-5^\circ,5^\circ)$. Figure \ref{fig:goal_positions} shows the possible constellations of the OS and the goal.

To generate the target vessels, we randomly sample a number of target ships from $\{0, 1, 2, 3\}$ with probabilities  $\{0.1, 0.3, 0.3, 0.3\}$ \rd{to develop} a robust and general final agent. \rd{Then}, we run the following COLREG-based routine for initiating a vessel:

\begin{itemize}
    \item First, we uniformly sample a COLREG situation $\sigma \in \{0, 1, 2, 3, 4\}$ and a corresponding heading intersection angle from the intervals in Table \ref{tab:COLREG_encounter_definitions}. If $\sigma = 0$, the null case, we randomly generate this angle from $\mathcal{U}(-67.5^\circ, 67.5^\circ)$, which constructs a ship with a \rd{course similar to that of} the OS. We found the inclusion of this case crucial since such cases do \rd{not otherwise} appear when a COLREG-dependent spawning procedure \rd{is used}. 
    \item Second, we sample a propeller movement from $\mathcal{U}(0.9, 1.1) \cdot 1.8$ revolutions per second, guaranteeing the existence of slower and faster ships \rd{compared to} the OS. Some prior \rd{researchers}, e.g. \cite{zhai2022intelligent}, \rd{initiated} all vessels at low or zero speed, which results in an acceleration phase with increased vessel \rd{manoeuvrability}. We avoid this simplifying procedure by solving the system of equations (\ref{eq:MMG_model_2}) for the longitudinal speed $u$ (fixing $v = r =0$) and \rd{initialise} the speed of the TS to the resulting value. The same procedure is \rd{used} for the OS. \rd{Furthermore}, when $\sigma = 4$ (overtaking), we multiply the derived speed of the TS by a realisation from $\mathcal{U}(0.3, 0.7)$, ensuring \rd{that the} vessel is slow enough to be overtaken.
    \item Third, \rd{with} the angle and velocity of the TS, we determine its position in $\{n\}$ by sampling a time $t_0$ from $\mathcal{U}(0.75, 1.0) \cdot \unit[25]{\rd{minutes}}$. The OS's advancement in the goal direction during $t_0$ is computed by determining the relative velocity of the OS towards the goal and forecasting this velocity over $t_0$. The procedure yields a point $(x_{n,t_0}, y_{n,t_0})$. The TS is initiated after time $t_0$ at $(x_{n,t_0}, y_{n,t_0})$, creating \rd{the need for} the OS to steer and avoid a collision. This is contrary to prior work, e.g. \rd{that of} \cite{xu2022path}, in which multiple target ships are randomly spawned in a simulation environment\rd{, resulting in} the possibility of not creating a threat to the OS.
\end{itemize}

\begin{figure}[h]
    \centering
    \includegraphics[width=\textwidth]{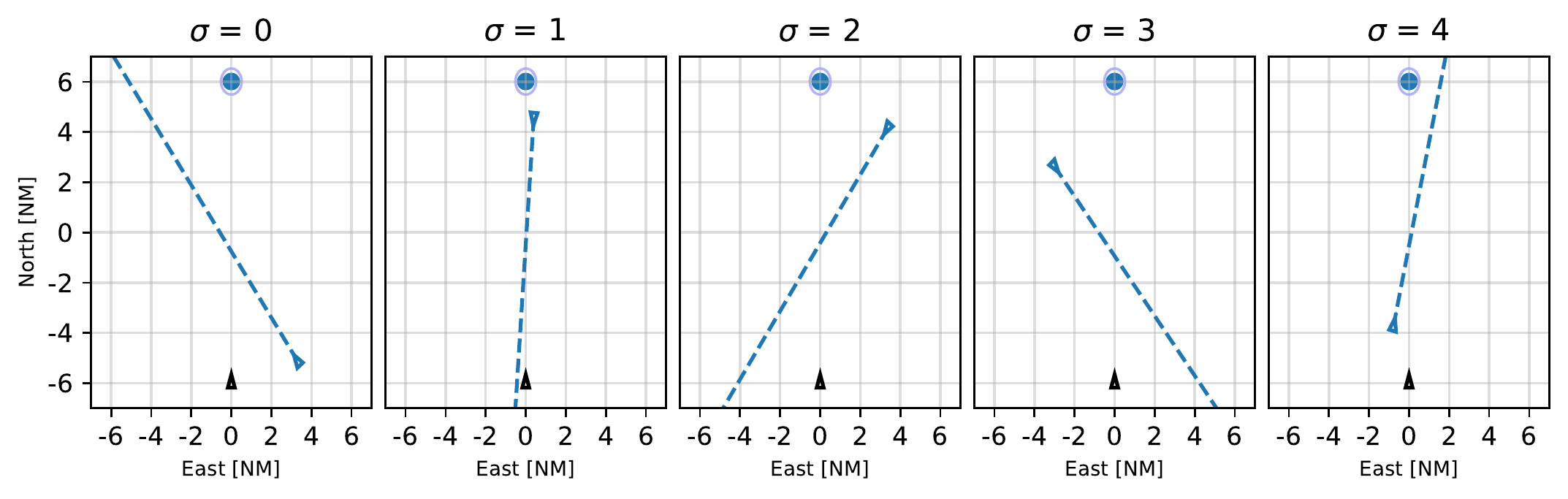}
    \caption{Sample of possible one-ship encounters during training.}
    \label{fig:TS_spawning}
\end{figure}

Figure \ref{fig:TS_spawning} \rd{visualises} exemplary scenarios in the case of one TS. An episode ends if the goal is reached or the number of episode steps is 1500. Crucially, we do not consider a collision an episode-ending event during training since we want to increase the number of highly negative reward transition tuples in the replay buffer. Recent research \rd{involving} other traffic simulations \citep{wurman2022outracing} shows that focusing on high-risk scenarios improves the robustness of the converged policy.

\subsection{Algorithm configuration and results}
\begin{figure}[ht]
    \centering
    \includegraphics[width=0.7\textwidth]{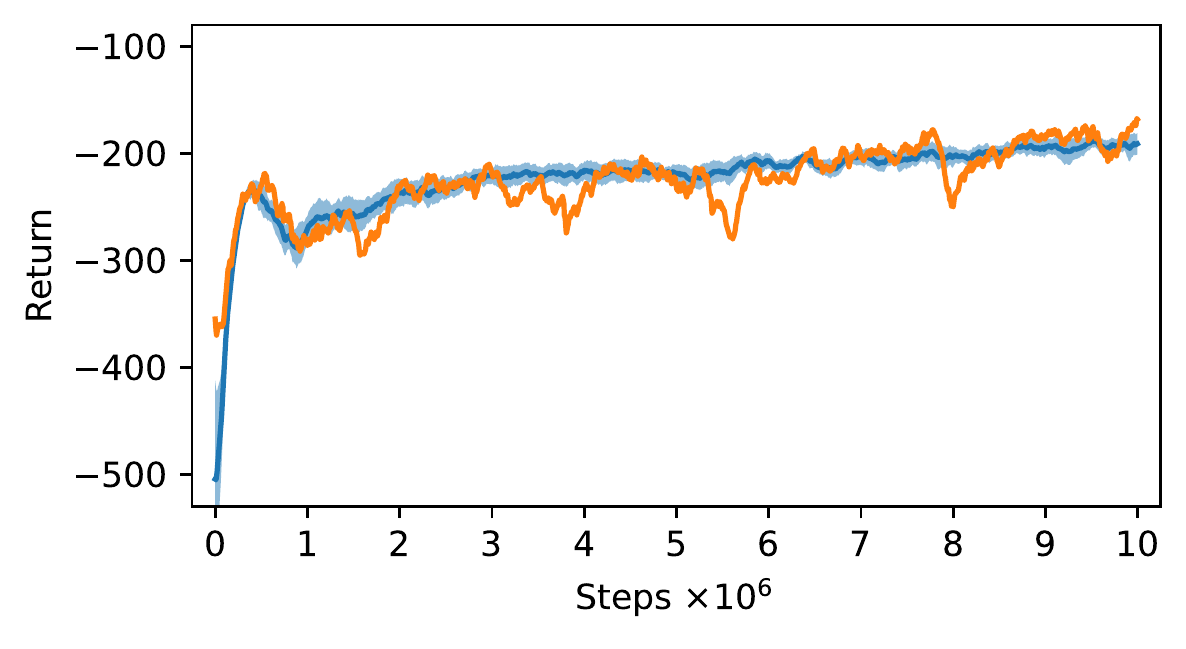}
    \caption{Test return over the training steps. The orange curve is the original run, which is used for validation. For comparison, we repeat the training and construct the blue curve, \rd{which represents} a mean over 15 independent seeds. The shaded area around the blue curve depicts 95\% point-wise confidence intervals and shows the algorithm's stability.}
    \label{fig:train_conf_int}
\end{figure}

We run the algorithm of \rd{subsection} \ref{subsec:spat_temp_arch} for $10^7$ steps. Table \ref{tbl:hyperparams} shows the hyperparameters, while the source code is accessible at \url{https://github.com/MarWaltz/TUD_RL}. During training, every 5,000 steps\rd{, we compute} the sum of rewards, called \rd{the} test return, of 10 evaluation episodes, average them, and exponentially smooth the results for clarity. Figure \ref{fig:train_conf_int} displays the training performance, \rd{which shows} a relatively stable improvement over the considered time steps.

\begin{table}[ht]
\footnotesize
    \def\arraystretch{1.1}
    \centering
    \begin{tabular}{ll}
    Hyperparameter & Value\\
    \toprule   
    Batch size & 32\\
    Discount factor ($\gamma$) &  0.999 \\
    Loss function & Mean squared error \\
    Min. replay buffer size & 1,000 \\
    Max. replay buffer size & $10^5$ \\
    \rd{Optimiser} & Adam \citep{kingma2014adam}\\
    Target network update frequency & $1,000$\\
    Initial exploration rate ($\epsilon_{\rm initial}$) & 1.0 \\
    Final exploration rate ($\epsilon_{\rm final}$) & 0.1 \\
    Test exploration rate ($\epsilon_{\rm test}$) & 0.0 \\
    Exploration steps & $10^6$ \\
    Time steps & $10^7$ \\
    History length ($h$) & 2 \\
    \end{tabular}
    \caption{List of hyperparameters. The $\epsilon$ \rd{value} of the $\epsilon$-greedy exploration scheme, which selects a random action with \rd{a} probability $\epsilon$ and \rd{selects} a greedy action otherwise, \rd{starts at} $1.0$ \rd{and} linearly \rd{decays} over $10^6$ steps to a final value of $0.1$.}
    \label{tbl:hyperparams}
\end{table}

\section{Validation}\label{sec:validation}
\subsection{\rd{Baselines}}
\rd{To assess the performance of our trained agent, we evaluate it in a range of scenarios involving single and multiple ships. To provide a benchmark for comparison, we choose two commonly used maritime path-planning techniques as baselines: the APF approach of \cite{lyu2019colregs} and the VO method of \cite{kuwata2013safe}. These works were selected because they are the seminal maritime implementations of the APF and VO methods, which are among the most widely applied techniques in the field, as recently reported by \cite{ozturk2022review}. Their corresponding hyperparameters are provided in \ref{appendix:APF_VO}.}

\rd{Our DRL approach combines local path planning and path following by controlling the rudder angle while considering the ship's dynamics. On the other hand, the APF and VO methods only perform local path planning and directly adjust the heading of the OS. These latter methods require a separate low-level control module to execute the planned trajectory, which can impact their performance in practice. Despite these differences, we include plots of the trajectories generated by all three approaches in the following figures. This allows for a comprehensive comparison of their planning abilities in the context of our study.}

\subsection{Around the Clock}\label{subsec:world}
\begin{figure}[htp]
    \centering
    \includegraphics[width=0.85\textwidth]{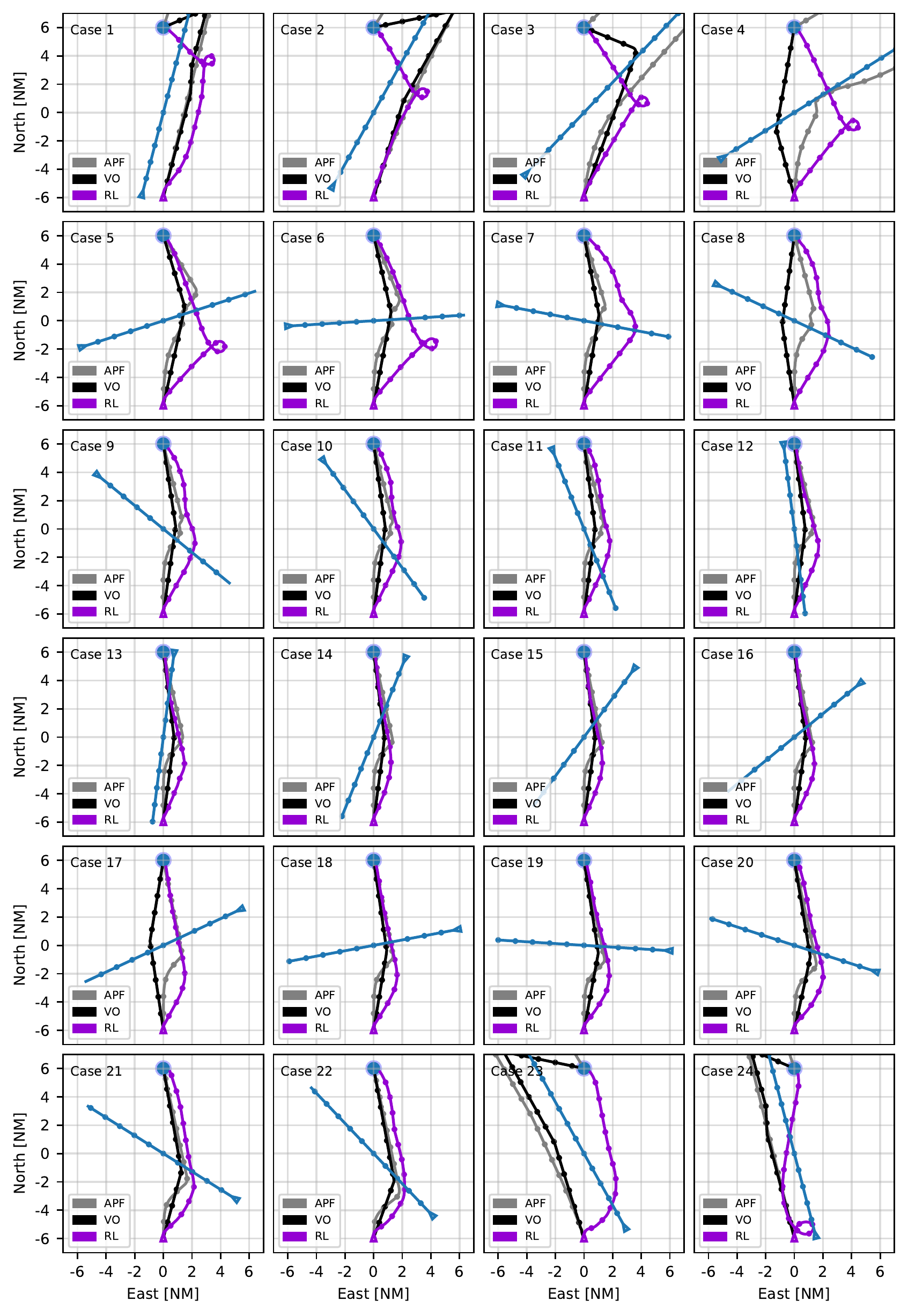}
    \caption{Around the Clock trajectories. The \rd{target ship} trajectories are \rd{blue}.~Two consecutive dots on a trajectory correspond to a 5\rd{-minute} time interval.}
    \label{fig:world_traj}
\end{figure}

\begin{figure}[htp]
    \centering
    \includegraphics[width=0.85\textwidth]{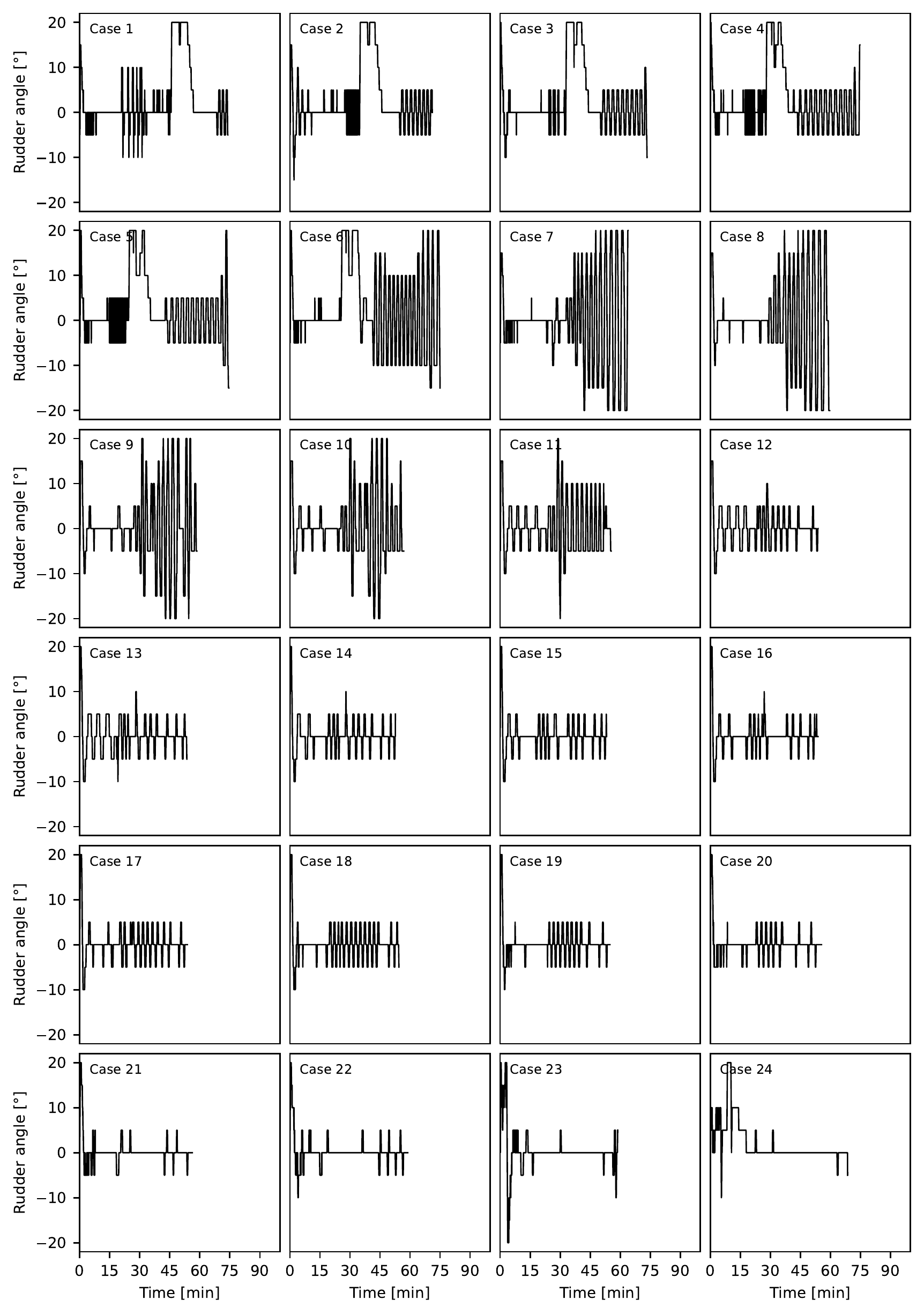}
    \caption{Rudder angles of the \rd{DRL agent} during the Around the Clock scenarios.}
    \label{fig:world_rudder}
\end{figure}

\rd{Regarding} single-ship situations, we create the 24 Around the Clock problems. These situations correspond to 24 equally spaced TS headings in the interval $(0, 2\pi)$. Precisely, we have $\phi_{TS,j} = \frac{j}{25} \cdot 2\pi$ for $j = 1, \ldots, 24$, where $j$ is the case number. Similar to \rd{subsection} \ref{subsec:env_setup}, we \rd{initialise} the OS and TS \rd{to be at} $o_n=(0,0)$ after $\unit[25]{\rd{minutes}}$ and place the goal on the opposite side of the simulation area. The revolutions per second \rd{value is equal to} 1.8 for all vessels. Figures \ref{fig:world_traj} and \ref{fig:world_rudder} show the trajectories and the OS's rudder angle, respectively, while the distances between the OS and the TS are depicted in Figure \ref{fig:world_dists} in \ref{appendix:Around_the_clock_distances}.

\rd{Upon analysing the trajectories generated by the DRL agent, we can observe that the agent successfully navigated to the goal while avoiding collisions with the linearly moving target ship. In all cases, the minimum encounter distance between the ships was greater than zero, demonstrating the effectiveness of the agent's COLAV strategy. Notably, in cases 1--6, the agent opted to execute a turning manoeuvre in order to maintain a safe distance from the other vessel, showcasing its ability to proactively avoid potential collisions. In later scenarios, the agent selected right-steering manoeuvres that were compliant with COLREG regulations, such as in case 16, when it avoided a starboard-crosser. In the final case, the agent made the interesting decision to immediately turn to the starboard side to allow the target ship to pass and then resumed its course towards the goal when the target ship was safely out of range.}

\rd{In all of the test scenarios, both the APF and VO approaches were able to successfully navigate without any collisions. However, the APF method exhibited some weaknesses in the first four cases; it suggested a path parallel to the target ship due to the superposition of attractive and repulsive forces. The VO approach performed better overall but still exhibited behaviour similar to that of the APF method in case 1, albeit to a lesser degree. One of the primary drawbacks of the VO method is the potential for bow-crossing with a target ship, which is seen in cases 3 and 6 and is generally considered to be unsafe in practice. In contrast, the DRL agent was able to avoid this behaviour entirely through the use of turning manoeuvres.}

\rd{In addition, the steering behaviour of the DRL agent is generally moderate, although there are some instances -- such as cases 7 and 8 -- in which the agent selects a sequence of alternating large positive and negative rudder angles. However, since each trajectory corresponds to over an hour of real time, these manoeuvres are still within the range of realistic behaviour. Notably, in case 24, the agent was able to maintain a steady course towards the goal for nearly half an hour without making any steering adjustments, demonstrating its ability to navigate successfully in the absence of any critical target ships.}

\subsection{Imazu problems}\label{subsec:Imazu}
\begin{figure}[htp]
    \centering
    \includegraphics[width=0.85\textwidth]{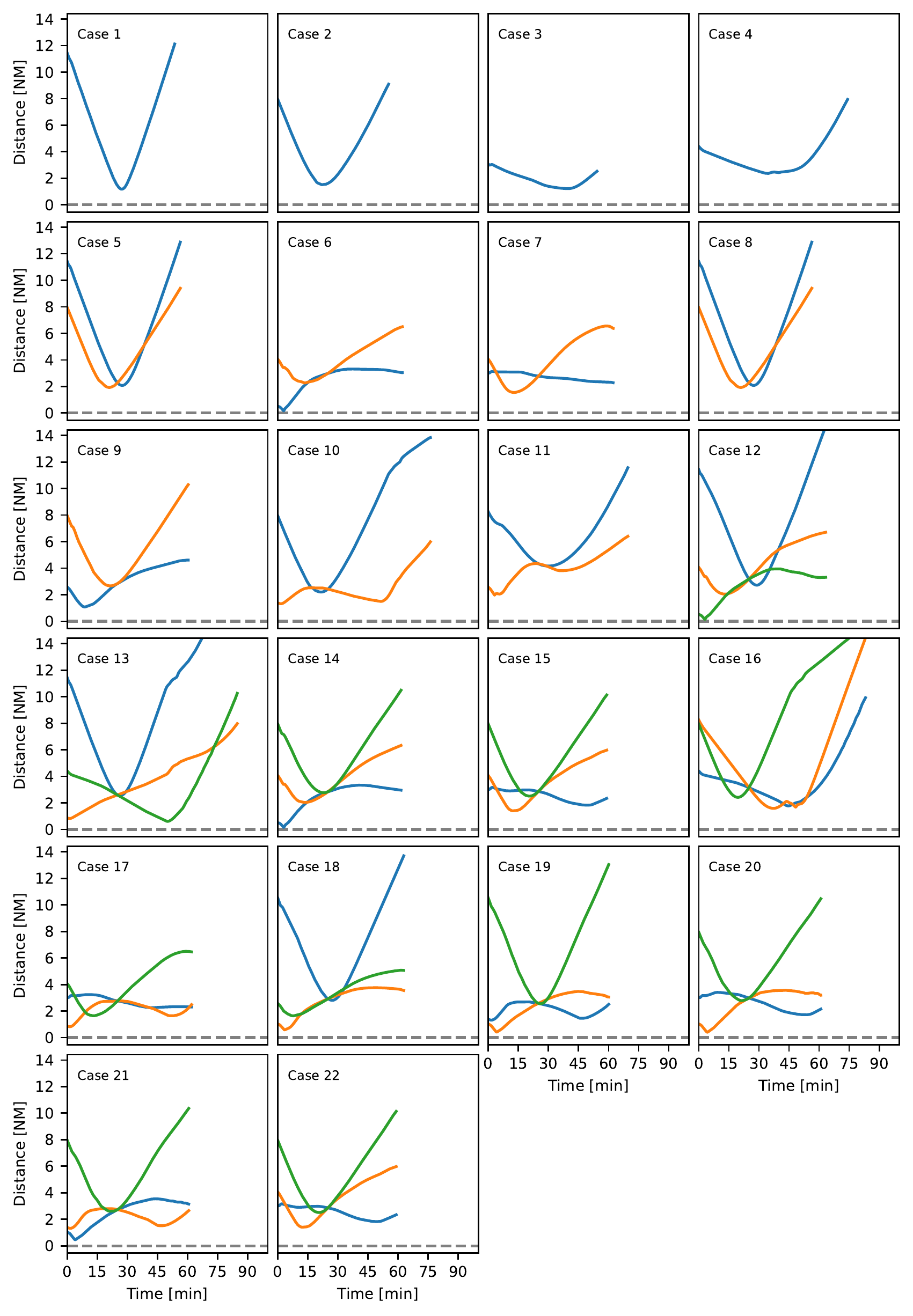}
    \caption{Distances of the \rd{DRL} agent to the target ships \rd{for} the Imazu problems. A \emph{distance of zero is not a physical collision}\rd{; it indicates} that the target ship is exactly at the border of the agent's ship domain.}
    \label{fig:Imazu_dist}
\end{figure}

\begin{figure}[htp]
    \centering
    \includegraphics[width=0.85\textwidth]{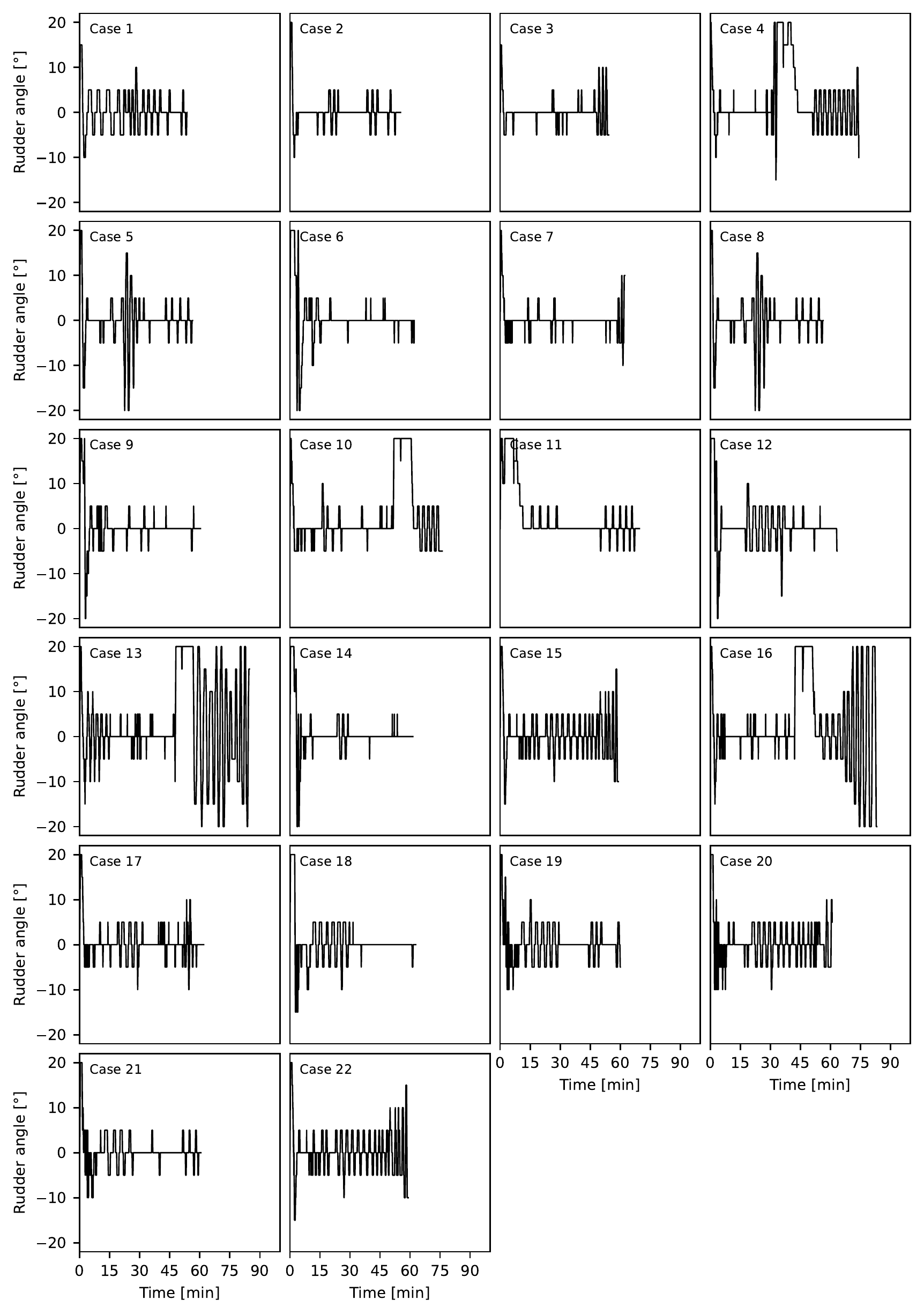}
    \caption{Steering \rd{behaviour} of the \rd{DRL} agent \rd{for} the Imazu problems.}
    \label{fig:Imazu_rudder}
\end{figure}

The second set of validation scenarios \rd{contains} the Imazu problems \citep{Imazu1987,sawada2021automatic, zhai2022intelligent}, a collection of single- and multi-ship encounters. The initial constellations of the target ships are detailed in Table \ref{tab:Imazu_definition} of \ref{appendix:Imazu_details}, while Figures \ref{fig:Imazu_traj}, \ref{fig:Imazu_dist}, and \ref{fig:Imazu_rudder} show the trajectories, \rd{the} distances between the OS and the target ships, and the rudder angles of the agent, respectively. Similar \rd{to} the Around the Clock scenarios, the \rd{DRL} agent smoothly reaches the desired goal area while successfully avoiding collisions with target ships \rd{for all problems}. Moreover, the COLREG compliance \rd{of the ship} is shown since the agent avoids head-on target ships and starboard-crossers \rd{by} steering to the right; see, e.g. cases 1, 2, 19, \rd{and} 22. Moreover, following rule 8 of the COLREGs, the agent takes \emph{substantial} \rd{COLAV} actions in these cases, which result in a sufficient perception of the agent's steering \rd{behaviour when there are} other ships in a situation at sea. Additionally, we do not observe any undesired bow-crossing of \rd{the} target ships, indicating the benefit of incorporating the factor $f_{\rm DCPA}$ in (\ref{eq:DCPA_prime}). 

An interesting characteristic of the \rd{DRL} agent's steering \rd{behaviour} is the frequent use of turning \rd{manoeuvres}. In the five cases 4, 10, 11, 13, and 16, the OS performs a starboard turn, which occurs at different times throughout the respective episode depending on the distance to the relevant TS. Such an advanced turning \rd{manoeuvre} was also observed \rd{by} \cite{sawada2021automatic}, whose agent performed a starboard turn only in case 4 of the same problem set. The authors attributed the ability to perform such a turning \rd{manoeuvre} to the specification of a continuous action space in the RL algorithm. However, due to our spatial-temporal recurrent neural architecture, we \rd{realise} an efficient information\rd{-}processing \rd{procedure} that still allows such \rd{a manoeuvre} with our simple, discrete action space. 

\rd{When comparing the three implemented methods, it is evident that the APF approach is the least effective at handling the encountered scenarios. As in the single-ship validations, the APF method often becomes trapped in a parallel situation next to a target ship, resulting in suboptimal paths and potentially unsafe situations. Additionally, the APF method struggles with symmetrical situations and can even lead to collisions in scenarios such as cases 19 and 21, where two closely positioned target ships are present from the start. On the other hand, the VO method performs much better and generally finds a short and collision-free path. However, as seen in cases 4, 6, and 13, it can still result in unsafe bow-crossing behaviour, which could pose a problem in real-life scenarios.}

\rd{Interestingly, the rudder movement plot (Figure \ref{fig:Imazu_rudder}) indicates that the DRL agent exhibits a low frequency of steering actions, with long phases with no steering at all, as seen in cases 11 and 18. Despite this, the agent still successfully navigated the complex scenarios presented in the validation, highlighting the effectiveness of the developed policy in handling real-world situations. Overall, the results demonstrate that our DRL approach offers advantages in terms of robustness and safety, and has the potential to be a valuable tool for practical path planning and COLAV applications in maritime environments.}

\begin{figure}[htp]
    \centering
    \includegraphics[width=0.85\textwidth]{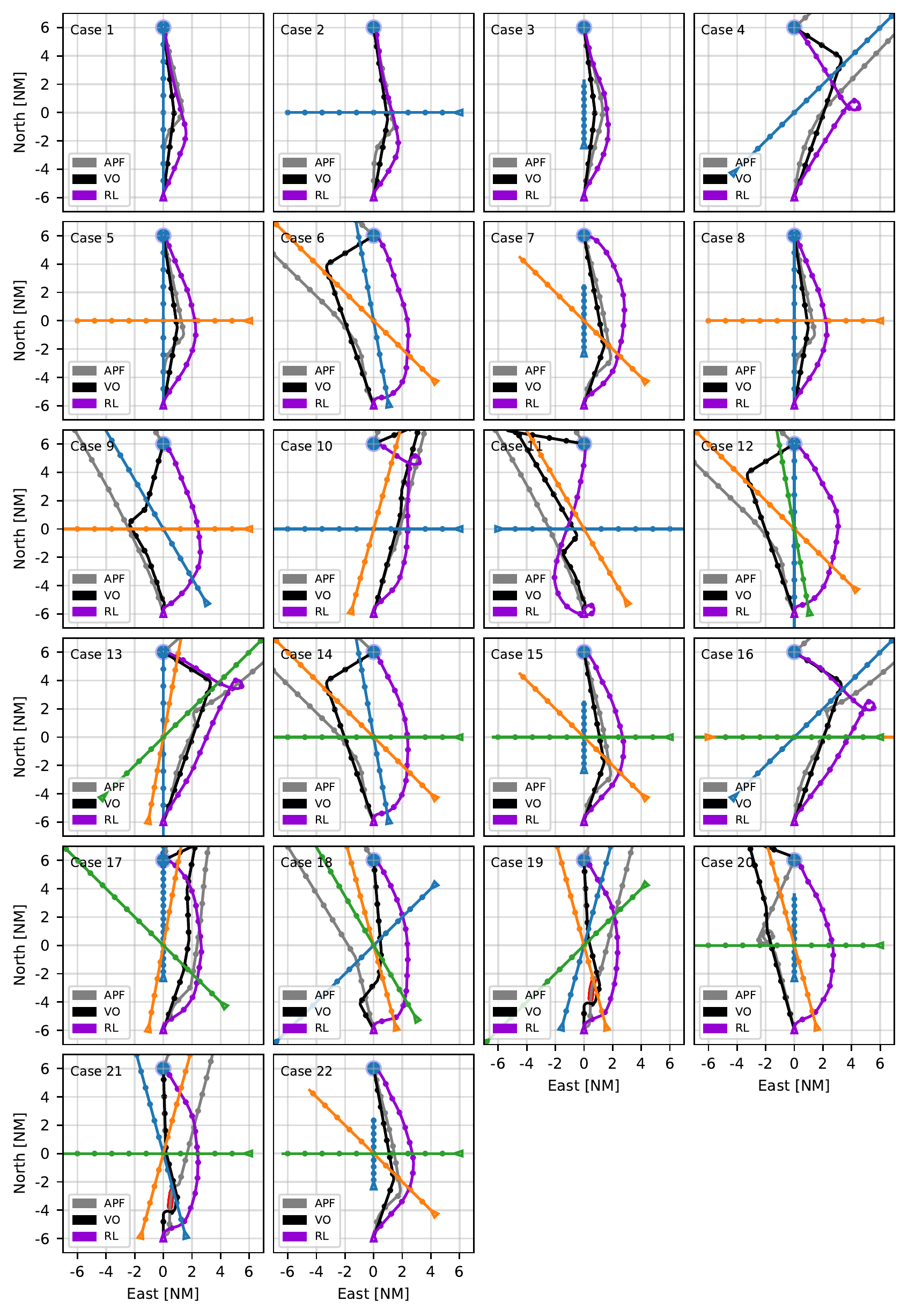}
    \caption{Trajectories \rd{for} the Imazu problems. The blue, orange, and green trajectories represent target ships\rd{, while a red area indicates a collision}.~Two consecutive dots on a trajectory correspond to a 5\rd{-minute} time interval.}
    \label{fig:Imazu_traj}
\end{figure}

\subsection{Star problems}
\begin{figure}[htp]
    \centering
    \subfloat{{\includegraphics[scale=0.6]{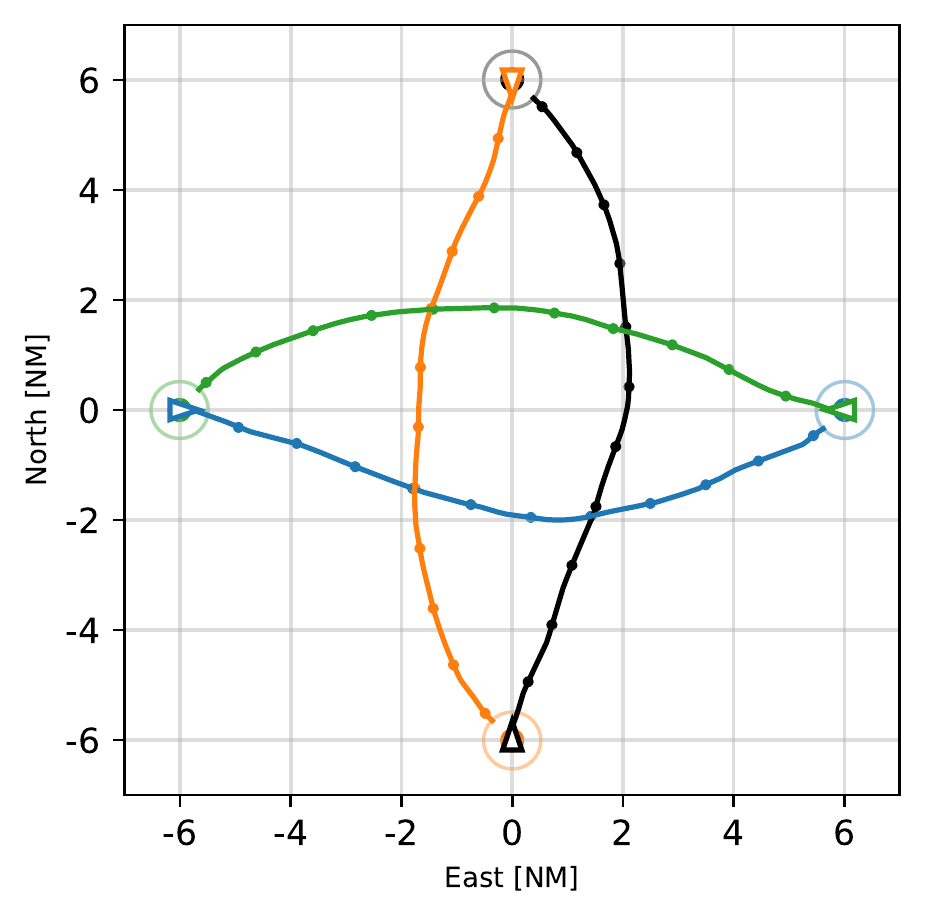} }}%
    \qquad
    \subfloat{{\includegraphics[scale=0.6]{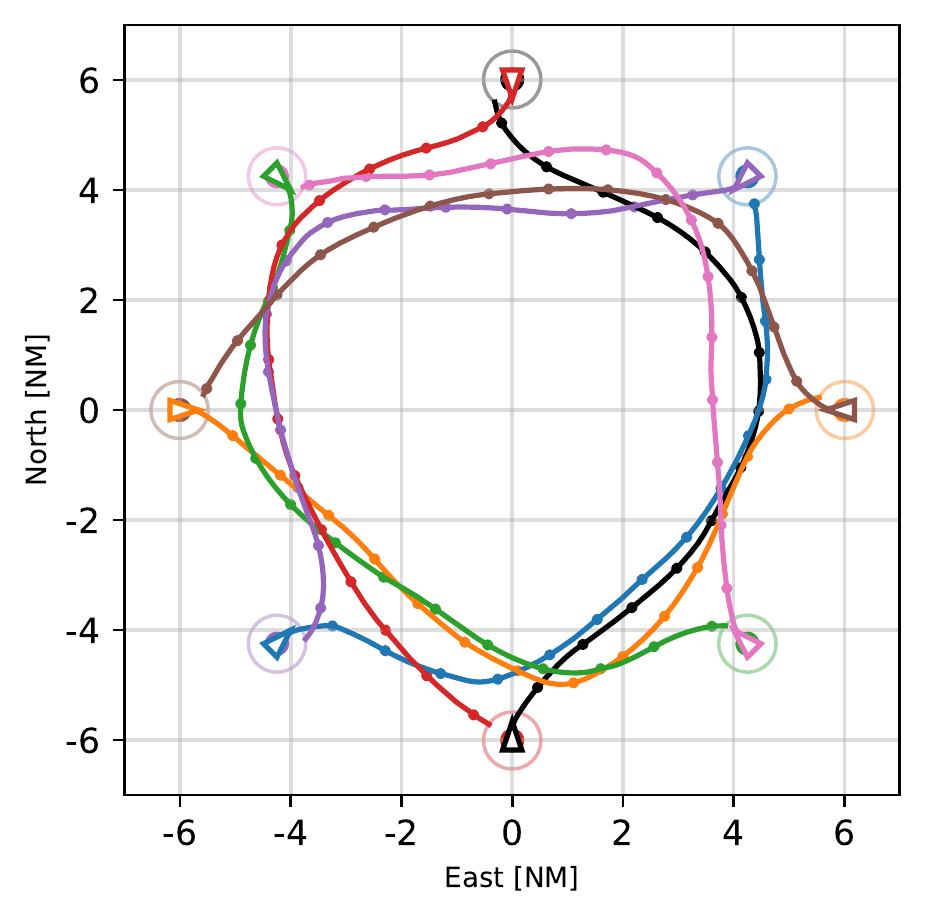} }}%
    \caption{Validation on the Star problems. Each ship represents \rd{a} DRL agent and follows the same policy. \rd{The} objective is to reach the opposite side \rd{of the grid}.}%
    \label{fig:Star_validation}%
\end{figure}

\rd{The training and validation scenarios presented so far have assumed a deterministic linear motion for the target ships, which is unrealistic in practice but common in the literature \citep{guo2020autonomous, fan2022novel, xu2022path}. To ensure that our final DRL policy can handle non-linear target ship movements, we tested it on two challenging multi-ship encounter scenarios known as Star problems \citep{zhao2019colregs}. The scenarios consist of four and eight ships, respectively, and the initial headings are spaced over $[0, 2\pi)$. Each ship is set to reach the origin $(0,0)$ after \unit[25]{minutes} if no steering takes place. Figure \ref{fig:Star_validation} shows the resulting trajectories, where each ship's goal is to reach the opposite ship's spawning area while following the same policy validated in previous subsections.}

\rd{Remarkably, the DRL policy successfully navigates through both scenarios, steering each ship to the right and then to the left to avoid collisions and reach the goal. In the eight-ship scenario, the COLAV actions are more intense, and the ships follow a more conservative, nearly circular path, reflecting the riskiness of a situation with so many vessels. These tasks are particularly challenging since the agent has never encountered more than three target ships in training nor was it ever exposed to non-linear moving ships. The policy's ability to handle such complex requirements demonstrates the neural architecture's flexibility and suitability for real-world maritime traffic applications.}

\section{Conclusion}\label{sec:conclusion}
\rd{Artificial intelligence has the potential to enhance safety, reduce accidents, and save energy resources in critical domains by either assisting or performing human actions. One such domain is maritime traffic, where technological advancements in ASV design coupled with regulatory adjustments could revolutionise how we perceive sea traffic. This study presents an ASV agent using deep Q-networks with a spatial-temporal recurrent neural network architecture to create a robust policy. The approach is validated on realistic multi-ship encounters, and it generates practically viable steering decisions aided by a newly proposed collision risk metric. In summary, our method has several advantages, including the ability to handle an arbitrary number of target ships, robustness to partial observability, state-of-the-art collision risk assessment, compliance with maritime traffic rules, and computational efficiency, making it feasible for practical deployment.}

\rd{Despite the success of our agent, there are limitations that we aim to address in future research. First, we assume that environmental disturbances such as wind, waves, and currents are not present. Incorporating these forces into our simulation is essential to ensure that our agent can perform adequately in real-world scenarios. Second, our agent relies on AIS data to obtain information about surrounding ships, and it is unaware of environmental characteristics such as the water depth, static obstacles, and non-AIS-equipped traffic participants. A future fully autonomous ASV should process additional sensor information and nautical maps to incorporate this information. Finally, all our work was performed in simulations, and we plan to conduct real-world experiments with miniature vessels to validate our architecture in the field.}

\section*{Acknowledgments}
\rd{The authors express their sincere gratitude to the anonymous reviewers for their valuable feedback, which significantly enhanced the quality of this manuscript. Additionally, the authors extend their thanks to Fabian Hart, Niklas Paulig, and Martin Treiber for their insightful comments and discussions. The authors also acknowledge} the Center for Information Services and High Performance Computing at TU Dresden for providing the resources for high-throughput calculations\rd{, which greatly supported this research.} This research did not receive any specific grant from funding agencies in the public, commercial, or not-for-profit sectors.


\bibliographystyle{apa}
\bibliography{bib}

\newpage
\appendix
\gdef\thesection{Appendix \Alph{section}}
\section{Selected COLREG rules}\label{appendix:COLREGs}
In the following, we present some of the rules from the \cite{COLREGs1972}.\\

\noindent\textbf{Rule 6: Safe speed}\\
\textit{Every vessel shall at all times proceed at a safe speed so that she can take proper and effective action to avoid collision and be stopped within a distance appropriate to the prevailing circumstances and conditions.}\\

\noindent\textbf{Rule 7: Risk of collision}\\
\textit{(a) Every vessel shall use all available means appropriate to the prevailing circumstances and conditions to determine if risk of collision exists. If there is any doubt such risk shall be deemed to exist.}\\

\noindent\textbf{Rule 8: Action to avoid collision}\\
\textit{(a) Any action to avoid collision shall be taken in accordance with the Rules of this Part and shall, if the circumstances of the case admit, be positive, made in ample time and with due regard to the observance of good seamanship.}\\

\noindent\textit{(b) Any alteration of course and/or speed to avoid collision shall, if the circumstances of the case admit, be large enough to be readily apparent to another vessel observing visually or by radar; a succession of small alterations of course and/or speed should be avoided.}\\

\noindent\textit{(c) If there is sufficient sea-room, alteration of course alone may be the most effective action to avoid a close-quarters situation provided that it is made in good time, is substantial and does not result in another close-quarters situation.}\\

\noindent\textit{(d) Action taken to avoid collision with another vessel shall be such as to result in passing at a safe distance. The effectiveness of the action shall be carefully checked until the other vessel is finally past and clear.}\\

\noindent\textbf{Rule 14: Head-on situation}\\
\noindent\textit{(a) When two power-driven vessels are meeting on reciprocal or nearly reciprocal courses so as to involve risk of collision each shall alter her course to starboard so that each shall pass on the port side of the other.}\\

\noindent\textbf{Rule 15: Crossing situation}\\
\noindent\textit{When two power-driven vessels are crossing so as to involve risk of collision, the vessel which has the other on her own starboard side shall keep out of the way and shall, if the circumstances of the case admit, avoid crossing ahead of the other vessel.}\\

\noindent\textbf{Rule 16: Action by give-way vessel}\\
\noindent\textit{Every vessel which is directed to keep out of the way of another vessel shall, so far as possible, take early and substantial action to keep well clear.}

\setcounter{table}{0}
\gdef\thesection{\color{red(ncs)} Appendix \Alph{section}}

\section{\rd{Hyperparameters of the baseline methods}}
\label{appendix:APF_VO}
\rd{In this section, we provide the hyperparameters of the baseline methods used in our experiments to ensure full reproducibility. We refer the reader to the respective papers for a detailed description of each method and the precise meaning of the parameters. Both methods generate output in the form of an OS heading, which we do not allow to deviate from the OS's current heading by more than $2.5^\circ$. This is to prevent these methods from having a significant advantage over the DRL approach in terms of manoeuvrability.}

\rd{The first baseline method is the APF approach proposed by \cite{lyu2019colregs}. To optimise the method's hyperparameters for our validation scenarios, we performed a small grid search and set the emergency scaling factor for close-range obstacles, $\eta_e$, to 5,000 and the safe distance, $d_{\rm safe}$, to 0.5 nautical miles. The remaining hyperparameters are the same as those used by \cite{lyu2019colregs}.}

\rd{The second baseline method is the VO approach outlined by \cite{kuwata2013safe}. Due to the lack of values in their paper, we replaced their COLREG situation classification with the values given in Table \ref{tab:COLREG_encounter_definitions} and optimised the remaining parameters via a small grid search. This resulted in the pre-collision check parameters being set to $t_{\max} = \unit[15]{minutes}$ and $d_{\min} = \unit[0.75]{NM}$, and the hysteresis parameter was set to $n_h = 60$. Since our DRL agent and the APF method of \cite{lyu2019colregs} only adjust the rudder angle and heading, respectively, we assume that the absolute value of the OS velocity is constant and only optimise for a heading angle in the VO method. At each step, we thus evaluate $N = 500$ equally spaced candidate headings over the interval $[\phi_{OS} - \frac{\pi}{2}, \phi_{OS} + \frac{\pi}{2}]$ and select the headings that are not in a VO- or COLREG-constrained velocity set (see \cite{kuwata2013safe}). We then select the heading that results in a velocity vector with a minimum 2-norm distance to the velocity vector towards the goal. This corresponds to setting the cost parameters of \cite{kuwata2013safe} to $\omega_{\tau} = 0$ and $\omega_v = 1$, and the weighting matrix $Q$ is the identity matrix.}

\setcounter{figure}{0}
\gdef\thesection{Appendix \color{red(ncs)}\Alph{section}}
\section{Around the \rd{Clock: Distances}}
\label{appendix:Around_the_clock_distances}
\begin{figure}[H]
    \centering
    \includegraphics[width=0.8\textwidth]{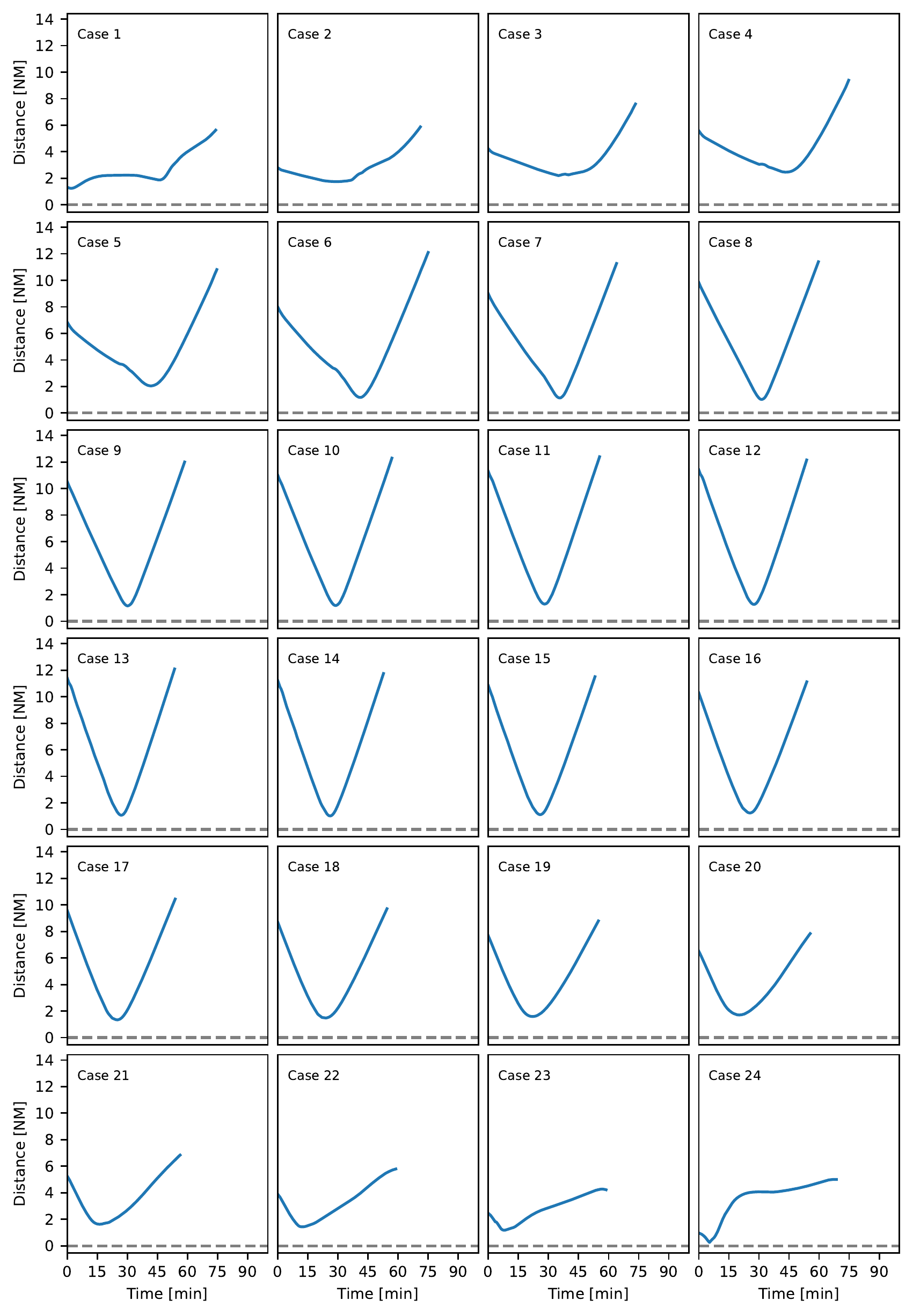}
    \caption{Distances between \rd{the DRL agent} and \rd{the target ship} during the Around the Clock scenarios. Following our definition of a collision, a distance of zero \rd{does} not equate to a physical collision\rd{; it indicates} that the TS's midship position is exactly at the border of the agent's ship domain.}
    \label{fig:world_dists}
\end{figure}

\setcounter{table}{0}
\gdef\thesection{Appendix \color{red(ncs)}\Alph{section}}
\section{Imazu problem constellations}
\label{appendix:Imazu_details}
\begin{table}[H]
\footnotesize
\def\arraystretch{1.1}
\setlength{\tabcolsep}{5.75pt}
\centering
    \begin{tabular}{r rrr rrr rrr}
\multirow{2}{*}{Case} & \multicolumn{3}{c}{Target Ship 1} & \multicolumn{3}{c}{Target Ship 2} & \multicolumn{3}{c}{Target Ship 3} \\
\cmidrule[0.05em]{2-10}
          & $\phi$ [$^\circ$] & N [NM] & E [NM] & $\phi$ [$^\circ$] & N [NM] & E [NM] & $\phi$ [$^\circ$] & N [NM] & E [NM]\\
         \toprule
         1 & 180 & 6.009 & 0.000 & -& -& - & - & - & -\\
         2 & -90 & 0.000 & 6.009 & - &- & -& - & - & -\\
         3 & 0 & -2.337 & 0.000 & - & -& -& - & - & -\\
         4 & 45 & -4.249 & -4.249 & - & -& -& - & - & -\\
         \midrule
         5 & 180 & 6.009 & 0.000 & -90 & 0.000 & 6.009 & - & - & -\\
         6 & -10 & -5.918 & 1.043 & -45 & -4.249 & 4.249 & - & - & -\\
         7 & 0 & -2.337 & 0.000 & -45 & -4.249 & 4.249 & - & - & -\\
         8 & 180 & 6.009 & 0.000 & -90 & 0.000 & 6.009 & - & - & -\\
         9 & -30 & -5.204 & 3.004 & -90 & 0.000 & 6.009 &  - & - & -\\
         10 & -90 & 0.000 & 6.009 & 15 & -5.804 & -1.555 &  - & - & -\\
         11 & 90 & 0.000 & -6.009 & -30 & -5.204 & 3.004 &  - & - & -\\
         \midrule
         12 & 180 & 6.009 & 0.000 & -45 & -4.249 & 4.249 & -10 & -5.918 & 1.043\\
         13 & 180 & 6.009 & 0.000 & 10 & -5.918 & -1.043 & 45 & -4.249 & -4.249\\
         14 & -10 & -5.918 & 1.043 & -45 & -4.249 & 4.249 & -90 & 0.000 & 6.009\\
         15 & 0 & -2.337 & 0.000 & -45 & -4.249 & 4.249 & -90 & 0.000 & 6.009\\
         16 & 45 & -4.249 & -4.249 & 90 & 0.000 & -6.009 & -90 & 0.000 & 6.009\\
         17 & 0 & -2.337 & 0.000 & 10 & -5.918 & -1.043 & -45 & -4.249 & 4.249\\
         18 & -135 & 4.249 & 4.249 & -15 & -5.804 & 1.555  & -30 & -5.204 & 3.004\\
         19 & 15 & -5.804 & -1.555 & -15 & -5.804 & 1.555  & -135 & 4.249 & 4.249 \\
         20 & 0 & -2.337 & 0.000 & -15 & -5.804 & 1.555  & -90 & 0.000 & 6.009\\
         21 & -15 & -5.804 & 1.555 & 15 & -5.804 & -1.555 & -90 & 0.000 & 6.009\\
         22 & 0 & -2.337 & 0.000 & -45 & -4.249 & 4.249 & -90 & 0.000 & 6.009\\
    \end{tabular}
    \caption{Starting positions of the target ships in the Imazu problems\rd{,} with heading angles \rd{built on the work of} \cite{sawada2021automatic} and \cite{zhai2022intelligent}.}
    \label{tab:Imazu_definition}
\end{table}






\end{document}